\documentclass[letterpaper]{article} 
\usepackage{aaai2026}  
\usepackage{times}  
\usepackage{helvet}  
\usepackage{courier}  
\usepackage[hyphens]{url}  
\usepackage{graphicx} 
\usepackage{natbib}  
\usepackage{caption} 
\usepackage{algorithm}
\usepackage{algorithmic}

\usepackage{newfloat}
\usepackage{listings}
\DeclareCaptionStyle{ruled}{labelfont=normalfont,labelsep=colon,strut=off} 
\floatstyle{ruled}
\newfloat{listing}{tb}{lst}{}
\floatname{listing}{Listing}
\usepackage{todonotes}

\usepackage{color,soul}
\usepackage{xcolor}
\definecolor{cerulean}{rgb}{0.0,0.48,0.65}
\definecolor{green}{rgb}{0.01, 0.75, 0.24}
\definecolor{Black}{RGB}{0.0, 0.0, 0.0}

\newcommand{\shadow}[1]{}
\def\s{\s}

\def\s{\shadow}

\usepackage{graphicx}
\usepackage{subfigure}
\usepackage{booktabs} 
\usepackage{tabularx}
\usepackage{multirow}
\usepackage{amsmath}
\usepackage{amssymb}
\usepackage{mathtools}
\usepackage{amsthm}
\usepackage{makecell}
\usepackage{adjustbox}

\title{MultiTab: A Scalable Foundation for Multitask Learning on Tabular Data}
\author{
    Dimitrios Sinodinos\textsuperscript{\rm 1,2},
    Jack Yi Wei\textsuperscript{\rm 1,2},
    Narges Armanfard\textsuperscript{\rm 1,2}
}
\affiliations{
    \textsuperscript{\rm 1}McGill University\\
    \textsuperscript{\rm 2}Mila - Quebec AI Institute\\
    dimitrios.sinodinos@mail.mcgill.ca, yi.wei4@mail.mcgill.ca, narges.armanfard@mcgill.ca
}

\usepackage{bibentry}

\begin{document}

\maketitle

\begin{tikzpicture}[remember picture, overlay]
    \node[anchor=north west, xshift=0.75in, yshift=-0.25in] at (current page.north west) {%
        \footnotesize Accepted for publication at AAAI 2026
    };
\end{tikzpicture}

\begin{abstract}
Tabular data is the most abundant data type in the world, powering systems in finance, healthcare, e‑commerce, and beyond. As tabular datasets grow and span multiple related targets, there is an increasing need to exploit shared task information for improved multitask generalization. Multitask learning (MTL) has emerged as a powerful way to improve generalization and efficiency, yet most existing work focuses narrowly on large‑scale recommendation systems, leaving its potential in broader tabular domains largely underexplored. Also, existing MTL approaches for tabular data predominantly rely on multi-layer perceptron-based backbones, which struggle to capture complex feature interactions and often fail to scale when data is abundant, a limitation that transformer architectures have overcome in other domains. Motivated by this, we introduce MultiTab-Net, the first multitask transformer architecture specifically designed for large tabular data. MultiTab-Net employs a novel multitask masked‑attention mechanism that dynamically models feature–feature dependencies while mitigating task competition. Through extensive experiments, we show that MultiTab-Net consistently achieves higher multitask gain than existing MTL architectures and single‑task transformers across diverse domains including large‑scale recommendation data, census‑like socioeconomic data, and physics datasets, spanning a wide range of task counts, task types, and feature modalities. In addition, we contribute MultiTab-Bench, a generalized multitask synthetic dataset generator that enables systematic evaluation of multitask dynamics by tuning task count, task correlations, and relative task complexity.
\end{abstract}

\begin{links}
    \link{Code}{https://github.com/Armanfard-Lab/MultiTab}
\end{links}

\section{Introduction}

Tabular data is a fundamental data format that supports a wide range of industries, including finance, healthcare, and e-commerce. Its structured layout of rows and columns enables efficient storage, querying, and manipulation, supporting critical applications from large-scale recommendation systems~\cite{peng2020improving} to population studies based on survey data and medical diagnosis ~\cite{uci_cdc_diabetes_2017}. As tabular data becomes increasingly ubiquitous, advancements in data collection and digital infrastructure have led to massive, high-dimensional datasets with multiple related prediction targets.  In healthcare, for instance, patient records containing age, BMI, and lab results can be used to predict both the presence of diabetes and the risk of high blood pressure. Similarly, on an online retail platform, it is desirable to not only predict if a user will click on a product, but also if they will add it to their cart and make a purchase \cite{su2024stem}. 

\begin{figure*}[htb!]
    \centering
    \includegraphics[width=0.96\linewidth]{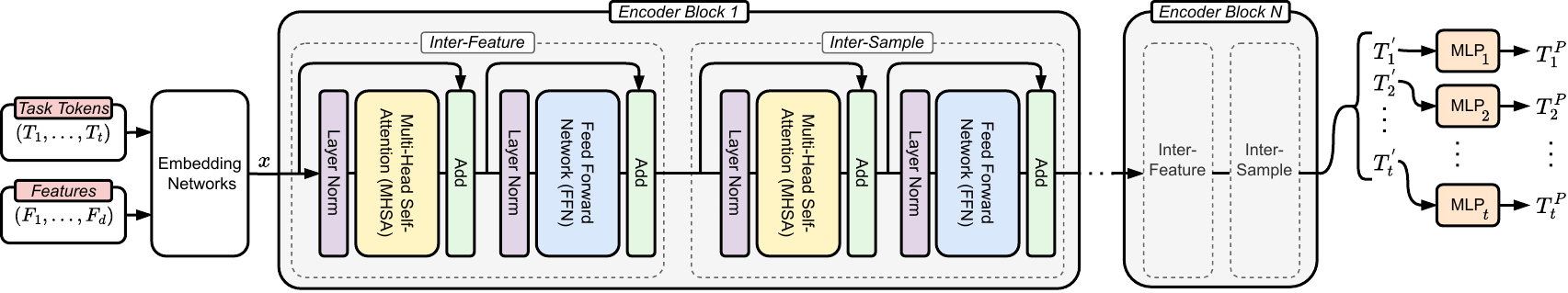}
    \caption{Overview of MultiTab-Net, our proposed multitask transformer for tabular data. A sample of $d$ features (categorical and/or numerical) and $t$ task tokens are passed through an embedding network to generate embeddings of size $e$ for each token. Task tokens are appended to the feature tokens, forming a combined input sequence, which is then processed by stacked encoder blocks. Each encoder block consists of inter-feature and inter-sample attention modules, along with feed-forward networks and residual connections. After $N$ encoder blocks, the processed task tokens $T' = \{T'_1, \dots, T'_t\}$ are passed through task-specific multilayer perceptrons (MLP) to generate the final task predictions $T^P = \{T^P_1, \dots, T^P_t\}$.}
    \label{fig:mtt}
\end{figure*}

In such settings, multitask learning (MTL)~\cite{caruana1997multitask} offers a way to leverage task correlations and improve generalization by training models on multiple tasks simultaneously. This shared learning enhances feature representations, mitigates overfitting, and acts as a form of regularization~\cite{vandenhende2021multi}. Beyond improving performance, MTL improves computational efficiency by enabling a single model to handle multiple tasks in parallel, reducing training and inference time~\cite{sinodinos2022attentive}.
In large-scale production environments, particularly recommendation systems, MTL has already proven effective in modeling related metrics such as click-through rate (CTR) and conversion rate (CVR) ~\cite{ma2018modeling, tang2020progressive, su2024stem}. However, MTL for tabular data remains largely underexplored outside of recommendation systems and existing approaches typically rely on simple multi-layer perceptron (MLP) backbones~\cite{ma2018modeling, tang2020progressive, su2024stem}. These MLP-based backbones lack explicit architectural mechanisms for modeling complex inter-feature interactions and task-specific dependencies ~\cite{chengWideDeepLearning2016}, which can limit their scalability as datasets and task counts grow~\cite{mcelfresh2023neural}.

Transformer architectures offer a promising alternative. While MLPs implicitly learn feature interactions through dense layers, transformers can leverage self-attention to dynamically model the relationships between features and tasks~\cite{vaswani2017attention}. In domains like NLP and vision, transformers have demonstrated their strength in capturing long-range dependencies and complex interactions, particularly when trained on large datasets ~\cite{devlin2018bert, dosovitskiy2020image}. Specifically for tabular data, transformers can also capture sample-level dependencies through self-attention, modeling relationships across rows and achieving substantial performance gains~\cite{somepalli2021saint}. This ability to model both inter-feature and inter-sample dependencies is particularly valuable for tabular datasets, where the relative importance of columns (features) and rows (samples) often varies across tasks. These advantages, combined with the increasing availability of large-scale, multi-target tabular datasets motivates the exploration of transformer-based architectures for MTL in tabular data.

We introduce MultiTab-Net, a novel multitask transformer designed specifically for tabular data. It combines the generalization and efficiency benefits of multitask learning with transformers' ability to model complex feature interactions via attention. A key innovation is our multitask masked attention framework, the first such mechanism applied to multitask transformers. While demonstrated on tabular data, this framework has broader potential in domains like computer vision and NLP, where managing task competition while capturing intricate feature interactions is equally critical. This mechanism prevents task interference in inter-feature learning~\cite{kendall2018multi}, leading to superior multitask gain, $\Delta_m$~\cite{maninis2019attentive}, on large-scale datasets, outperforming traditional multitask MLPs and single-task transformers. Notably, the magnitude of these gains exceeds those typically reported in prior multitask tabular works~\cite{ma2018modeling,tang2020progressive,su2024stem}.

In addition to public tabular datasets, synthetic data has become a standard approach for evaluating multitask learning performance and robustness to variations in task correlation~\cite{ma2018modeling, tang2020progressive}. The existing method by~\cite{ma2018modeling} allows for tunable task correlations but is limited to two tasks and lacks a notion of relative task difficulty—both key factors in multitask learning dynamics~\cite{vandenhende2021multi}. These constraints limit its effectiveness in analyzing multitask behaviors. To address this, we introduce MultiTab-Bench, a novel synthetic multitask dataset generator that enables full tunability of pairwise task correlations and relative task difficulty for any number of tasks, providing a more comprehensive synthetic baseline for multitask evaluations in tabular settings.

Overall, our contributions are as follows:
\begin{itemize}
    \item We introduce \textbf{MultiTab-Net}, the first multitask transformer architecture for tabular data, which outperforms existing MLP-based multitask models and single-task transformer models on widely used tabular benchmarks.
    \item We propose a \textbf{novel multitask masked attention} method that mitigates the effects of task competition during inter-feature relationship learning.
    \item We develop \textbf{MultiTab-Bench}, a new synthetic multitask tabular dataset generator that enables fine-grained control over pairwise task correlations and relative task difficulty for any number of tasks, allowing for evaluation on virtually any multitask learning setting.
\end{itemize}


\section{Related Works}
\label{related works}

\textbf{Deep Learning} has transformed domains like vision and language~\cite{krizhevskyImageNetClassificationDeep2012, devlin2018bert}, but has historically struggled to handle the heterogeneous feature types and irregular distributions common in tabular data. Recent studies now show that deep learning can also achieve competitive or superior performance on tabular benchmarks~\cite{mcelfresh2023neural, ericksonTabArenaLivingBenchmark2025a}. Multi-Layer Perceptrons (MLPs) remain strong baselines when paired with careful regularization, tuning, and embedding strategies~\cite{kadra2021well, gorishniy2022embeddings}. Building on this, transformer-based architectures such as TabTransformer (TabT)~\cite{huang2020tabtransformer}, FT-Transformer (FT-T)~\cite{gorishniy2021revisiting}, and SAINT~\cite{somepalli2021saint} adapt attention mechanisms to tabular data, yielding consistent gains. More recently, foundation models like TabPFN~\cite{hollmann2022tabpfn, hollmann2025accurate} and TabICL~\cite{qu2025tabicl} introduced in-context learning for tabular settings, but their scalability is limited to datasets with fewer than 100k samples, making them unsuitable for large-scale applications.

\textbf{Multitask learning} in tabular data has received limited attention outside of recommendation systems. MMoE~\cite{ma2018modeling} introduced a multitask mixture-of-experts (MoE) architecture using MLPs. PLE~\cite{tang2020progressive} extended this design by partitioning experts into shared and task-specific groups, mitigating task interference through a more structured gating scheme. STEM~\cite{su2024stem} then incorporated task-specific and shared embeddings and introduced stop-gradient constraints to prevent cross-task updates during backpropagation. This idea of controlling cross-task interactions directly influenced the multitask masked attention mechanism in MultiTab-Net. To the best of our knowledge, MultiTab-Net is the first transformer-based MTL model for tabular data, offering a scalable architecture and novel attention design tailored for multitask learning.

\section{MultiTab-Net: A Multitask Tabular Transformer}
\label{method}
We propose MultiTab-Net, a novel multitask transformer architecture designed for tabular data. It introduces two major innovations: (1) a multi-token mechanism that enables parallel task processing while capturing both shared and task-specific information, and (2) a multitask masked attention mechanism that encourages task-specific focus and inhibits task interference for inter-feature attention.

\subsection{MultiTab-Net Architecture}
The architecture of our multitask transformer, illustrated in Figure~\ref{fig:mtt}, builds upon existing transformer-based tabular models, such as FT-Transformer~\cite{gorishniy2021revisiting} and SAINT~\cite{somepalli2021saint}, that adopt the BERT-style approach~\cite{devlin2018bert} by adding an input token embedding corresponding to the downstream task. We extend this idea to a multitask setting by introducing a distinct task token $T_i$ for each of the $t$ tasks. This multi-token design enables parallel task processing, allowing each task token to interact with shared feature representations while retaining task-specific information. In Table \ref{tab:ablation}, we demonstrate empirically that this multi-token approach, combined with appropriate masking, outperforms the na\"ive multitask variant of using one shared task token as input to $t$ decoder heads. 

The model first processes the $t$ task tokens and $d$ input features (both categorical and numerical) through separate embedding networks for each token. Following~\cite{gorishniy2021revisiting, somepalli2021saint, su2024stem}, the embedding networks ensure that all tokens are transformed into appropriate vector representations with embedding size $e$ for the transformer layers. The resulting tokens are concatenated to form a sample $x \in \mathbb{R}^{(d+t)\times e}$, which is fed into the encoder blocks. Following~\cite{somepalli2021saint}, each encoder block contains two types of attention layers: Inter-Feature and Inter-Sample. The Inter-Feature layer models relationships across features (columns) within a sample. This mechanism is central to all transformers. In our model, this layer also allows tasks to attend to features and vice versa. The Inter-Sample attention layer captures patterns between samples (rows) in a batch, which has been shown to improve generalization for tabular data~\cite{somepalli2021saint}. Each attention layer is followed by a feed-forward network (FFN), and layer norms are used to improve training stability. After $N$ encoder blocks, the task tokens are sent to task-specific MLPs to generate each task prediction, $T_i^P$.

\subsection{Multitask Masked Attention}
\label{MMA}
In extending transformers to a multitask setting for tabular learning, the introduction of task tokens fundamentally changes the structure of the inter-feature (column-wise) attention. Instead of attention operating solely between feature tokens, the attention matrix now contains additional quadrants; where task tokens can attend to other task tokens, task tokens can attend to feature tokens, and feature tokens can attend to task tokens. This richer interaction space enables task tokens to explicitly influence other task and feature tokens. However, it also introduces new risks of task competition and the ``seesaw phenomenon''~\cite{tang2020progressive}, where certain tasks dominate the shared capacity and harm overall performance~\cite{vandenhende2021multi}. Since attention scores directly govern how tokens influence each other, they present an intuitive point of intervention. If task tokens are allowed to freely influence each other or the feature tokens, dominant tasks may distort attention distributions, biasing feature–feature interactions and harming generalization. To address this, we investigate explicit masking strategies that selectively remove certain connections in the attention map, thereby limiting the pathways through which interference can arise. Concretely, we propose and evaluate three candidate masking schemes, each grounded in logical intuition:
\begin{itemize}
    \item \textbf{F$\not\to$T}: Attention scores for the task tokens are masked from the feature tokens. The idea is to have only feature tokens influence other feature tokens, without influence from task tokens that may compete for representation.
    \item \textbf{T$\not\to$T}: Attention scores between different task tokens are masked. The idea is to limit their direct influence over each other and to further mitigate task competition.
    \item \textbf{F$\not\to$T \& T$\not\to$T}: Attention scores for the task tokens are masked from the feature tokens and different task tokens. The idea is to compound the benefits from both schemes.
\end{itemize}

We now present the mathematical formulation of our masking method in the context of Inter-Feature attention.
Given a sample $x \in \mathbb{R}^{(d+t) \times e}$ obtained by appending $t$ task tokens to $d$ feature tokens (each with embedding size $e$), we compute the multi-head attention scores for the $i^{\text{th}}$ head and the output $x_{\text{out}}$ as follows:

\begin{equation}
Q_i = x W_{Q_i}, \quad K_i = x W_{K_i}, \quad V_i = xW_{V_i},
\label{eq:qkv}
\end{equation}
\begin{equation}
A_i = \text{softmax}\left(\frac{Q_i K_i^\top}{\sqrt{d_k}}\right), \quad x_{\text{att},i} = A_i V_i
\end{equation}
\begin{equation}
    x_{\text{out}} = \text{Concat}(x_{\text{att},1}, x_{\text{att},2}, ..., x_{\text{att},h})W_O
\end{equation}
Here, $W_{Q_i}, W_{K_i}, W_{V_i} \in \mathbb{R}^{e \times d_k}$ are learnable projection matrices, $d_k$ is the attention head dimension split across $h$ heads such that $d_k = \frac{e}{h}$ for each head, and $A_i \in \mathbb{R}^{(d + t) \times (d + t)}$ captures the attention scores among all tokens. The outputs from all heads are concatenated and multiplied with the output projection layer $W_O \in \mathbb{R}^{e \times e}$ to form $x_\text{out}$. To prevent any task from disproportionately influencing the attention process, we apply a mask $M_A$ to selectively constrain the attention maps $A_i$ according to Equation ~\ref{eq:mask_apply}.

\begin{figure}
    \centering    \includegraphics[width=0.95\linewidth]{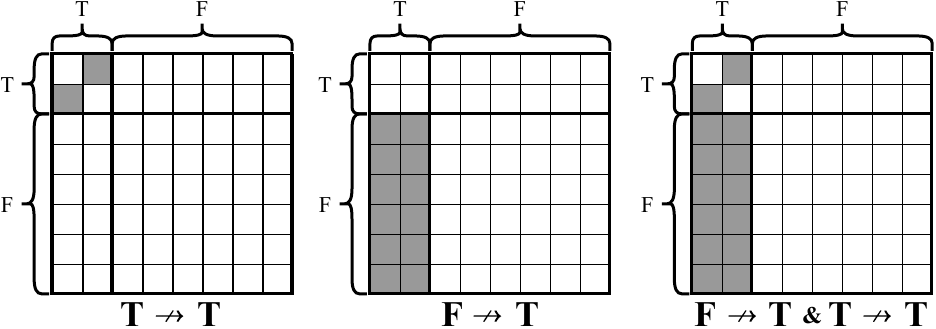}
    \caption{Illustration of the attention mask $M_A$ under different masking schemes. Masked cells are shaded in grey and unmasked in white. In this example, there are two task tokens and six feature tokens. In \textbf{T $\not\to$ T}, task tokens do not attend to other task tokens. Similarly, in \textbf{F $\not\to$ T}, feature tokens do not attend to task tokens. Finally, \textbf{F $\not\to$ T \& T $\not\to$ T} combines both schemes.}
    \label{fig:masks}
\end{figure}

\begin{equation}
\begin{aligned}
    A_i &= \text{softmax}\left(\frac{Q_i K_i^\top}{\sqrt{d_k}}+M_A\right),\\
\end{aligned}
\label{eq:mask_apply}
\end{equation}

In practice, the mask $M_A$ modifies the pre-activation attention scores by adding $-\infty$ to masked positions and 0 to unmasked ones. After applying the softmax activation, the masked entries become 0, and the unmasked scores are unchanged. The candidate masking schemes are illustrated in Figure~\ref{fig:masks}. It is worth noting that we do not evaluate masking the feature tokens from the task tokens (i.e., \textbf{T$\not\to$ F}), as task tokens must attend to feature tokens for task-specific information extraction, as shown across numerous domains ~\cite{dosovitskiy2020image, somepalli2021saint, devlin2018bert}. In contrast to inter‑feature attention, which relies on explicit token‑to‑token interactions that can be masked, inter‑sample attention pools information across entire samples, making masking inapplicable. The formulation of our multitask inter‑sample attention is detailed in the Appendix.

\section{MultiTab-Bench:  A Generalized Synthetic Multitask Dataset Generator}

In tabular learning, synthetic data generation has become a valuable tool for evaluating supervised learning performance~\cite{ma2018modeling} and even for achieving impressive zero-shot performance on real data when generated with sufficient diversity in underlying distributions and feature interactions~\cite{hollmann2022tabpfn, hollmann2025accurate}. However, unlike domains such as computer vision~\cite{zamir2018taskonomy, mottaghi_cvpr14}, there is a notable lack of evaluations focused on multitask tabular learning involving more than two tasks. This represents an important gap in the literature, as it has been shown that effective multitask learning models are capable of managing complex training dynamics associated with an increasing number of tasks~\cite{standley2020tasks, 10943529}. To address this, we introduce MultiTab-Bench, a novel synthetic dataset generator designed to evaluate multitask performance across an arbitrary number of tasks with tunable pairwise task correlations and relative task complexity, offering ample flexibility for simulating diverse multitask settings.

\subsection{Formulation of Weight Vectors}\citeauthor{ma2018modeling} (2018) introduced a synthetic multitask dataset generator designed for exactly two tasks. They would generate a weight vector for each task, $w_1$ and $w_2$, such that their cosine similarity matched a predefined correlation constant $p$. These weight vectors were applied to samples drawn from a standard normal distribution through a series of non-linear transformations, which empirically demonstrated a predictable influence on the Pearson correlation between the output labels. While this setup allowed for flexible control over task correlations, it was limited to two tasks and lacked the complexity needed to assess the robustness of modern multitask models. Building on their approach, we propose a generalized multitask synthetic dataset generator that supports full tunability of task correlations for an arbitrary number of tasks, while introducing additional complexities such as relative task difficulty and task-specific uncertainty~\cite{kendall2018multi}. Specifically, we formulate our task-specific weight vectors as follows:
\begin{enumerate}
\item \textbf{Define the Correlation Matrix} $\mathbf{P}$: We define $\mathbf{P} \in \mathbb{R}^{t \times t}$ as a symmetric correlation matrix where $P_{ij} = 1$ if $i = j$ to represent self-correlation, and $P_{ij} = p$ if $i \neq j$ to represent the desired pairwise correlation between tasks.

\item \textbf{Perform eigendecomposition of} \( \mathbf{P} \): Since \(\mathbf{P}\) is real symmetric, we have \( \mathbf{P} = \mathbf{Q} \boldsymbol{\Lambda} \mathbf{Q}^T \), where $\mathbf{Q} \in \mathbb{R}^{t \times t}$ is the orthogonal matrix of eigenvectors (not to be confused with the query matrix used in attention), and $\boldsymbol{\Lambda} \in \mathbb{R}^{t \times t}$ is the diagonal matrix of eigenvalues $\lambda_1, \lambda_2, \dots, \lambda_t$. By definition of eigendecomposition, the matrix entries $P_{ij}$ can be expressed as $P_{ij} = \sum_{k=1}^t Q_{ik} Q_{jk} \lambda_k$, where $Q_{ik}$ and $Q_{jk}$ denote the entries of matrix $\mathbf{Q}$ corresponding to the $i^{\text{th}}$ and $j^{\text{th}}$ rows, respectively, and the $k^{\text{th}}$ column. Since $0 \leq p \leq 1$ and $t \geq 2$, $\mathbf{P}$ is guaranteed to be positive semidefinite (see Horn \& Johnson, 2012) and thus the entries of \(\boldsymbol{\Lambda}\) are non-negative. 

\item \textbf{Construct the task-specific weight matrix}: Expressed as \( \mathbf{W} = \mathbf{Q} \boldsymbol{\Lambda}^{1/2} \mathbf{U}^T \),
where \( \boldsymbol{\Lambda}^{1/2} \) is the diagonal matrix containing the square roots of the eigenvalues, \( \mathbf{U} \in \mathbb{R}^{d \times t} \) is a matrix with columns of orthogonal unit vectors \( \mathbf{u}_1, \mathbf{u}_2, \dots, \mathbf{u}_t \in \mathbb{R}^{d} \).
\end{enumerate}

We can now verify cosine similarity between weight vectors \( \mathbf{w}_i \) and \( \mathbf{w}_j \): \( \text{cosine\_similarity}(\mathbf{w}_i, \mathbf{w}_j) = \smash{\frac{\mathbf{w}_i^T \mathbf{w}_j}{\|\mathbf{w}_i\| \|\mathbf{w}_j\|}} \). Let \(\mathbf{w}_i\) be the $i^{\text{th}}$ row of \(\mathbf{W}\), from \( \mathbf{w}_i  = \sum_{k=1}^t Q_{ik} \sqrt{\lambda_k} \mathbf{u}_k \) and the orthogonality of \( \mathbf{U}\), we derive: \( \mathbf{w}_i^T \mathbf{w}_j = \sum_{k=1}^t Q_{ik} Q_{jk} \lambda_k = P_{ij} \). Since \( \|\mathbf{w}_i\| = 1 \), the cosine similarity is exactly \( P_{ij} \). Further details about the intermediate steps in the proof can be found in the appendix.

\subsection{Data Generation}
To generate task labels with predictable pairwise Pearson correlations, we first produce task-specific weight vectors with tunable pairwise cosine similarity. The data generation process proceeds as follows:
\begin{enumerate}
    \item \textbf{Generate Input Features:}  
    Similar to the approach in~\cite{ma2018modeling}, the input features \( \mathbf{x} \in \mathbb{R}^d \) are sampled from a standard multivariate normal distribution:  \( \mathbf{x} \sim \mathcal{N}(0, \mathbf{I}) \)
    \item \textbf{Generate Labels:}  
    Task-specific labels for the $i^{\text{th}}$ task are then generated using a polynomial transformation combined with task-specific noise such that \( y_i = \sum_{k=1}^{d_i} ( \mathbf{w}_i^T \mathbf{x})^k + \epsilon_i \), and \( \epsilon_i \sim \mathcal{N}(0, \sigma_i^2) \), where \( d_i \) is the polynomial degree and \( \sigma_i^2 \) controls the task-specific noise variance, introducing uncertainty into the labels.
\end{enumerate}
Prior work~\cite{kendall2018multi} highlights the link between task uncertainty, difficulty, and training balance. By adding task-specific noise, our method enables finer control over relative task difficulty.

\begin{figure}[htb!]
    \centering
    \includegraphics[width=0.97\linewidth]{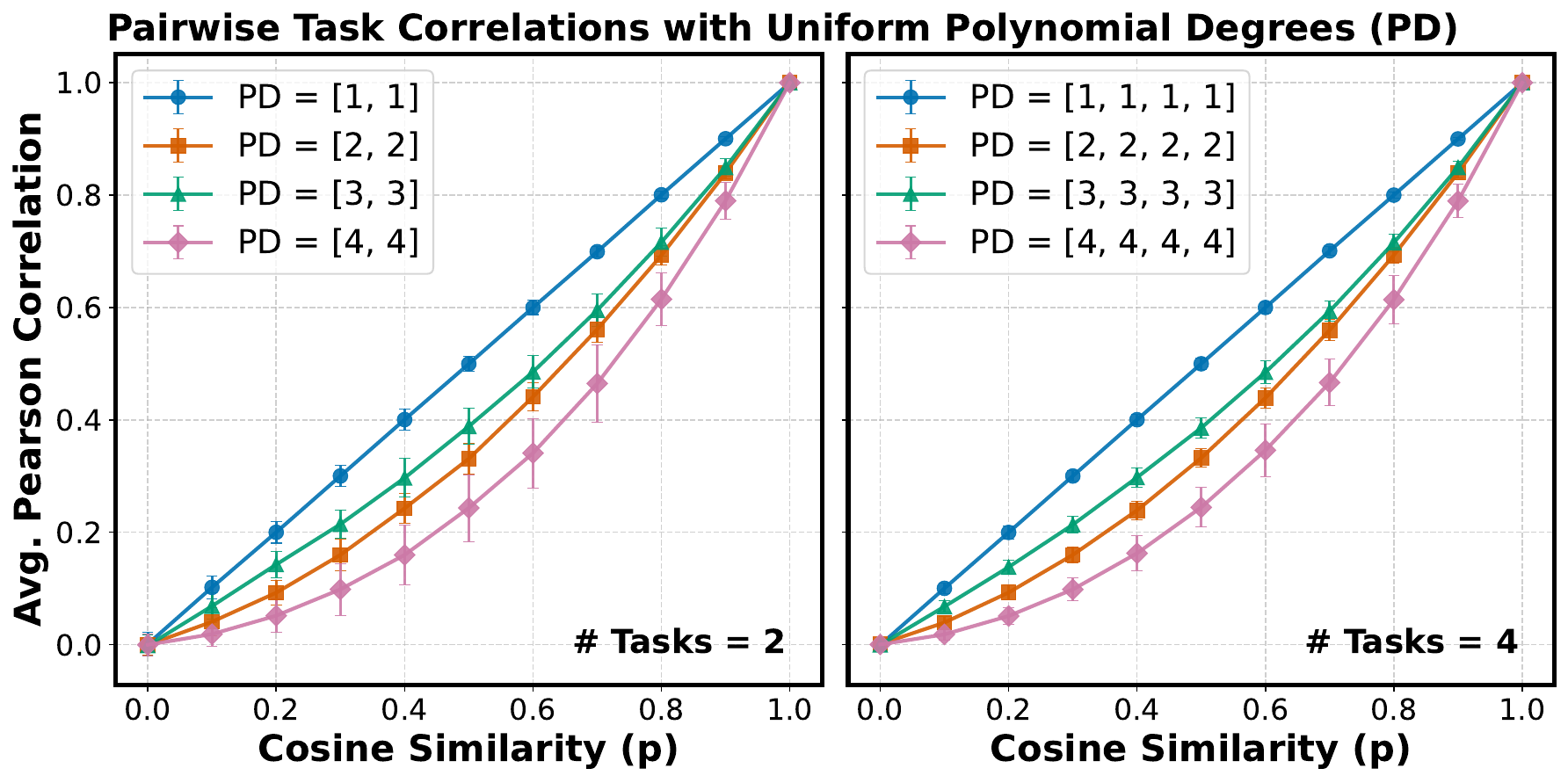}
    \caption{Average pairwise Pearson correlation for two and four tasks using the same polynomial degree for each task label. PD = [1, 1] indicates that two tasks were generated using a polynomial of degree 1, and similarly for other PD values and task counts.
}
    \label{fig:uniform}
\end{figure}

\begin{figure}[htb!]
    \centering
    \includegraphics[width=0.9\linewidth]{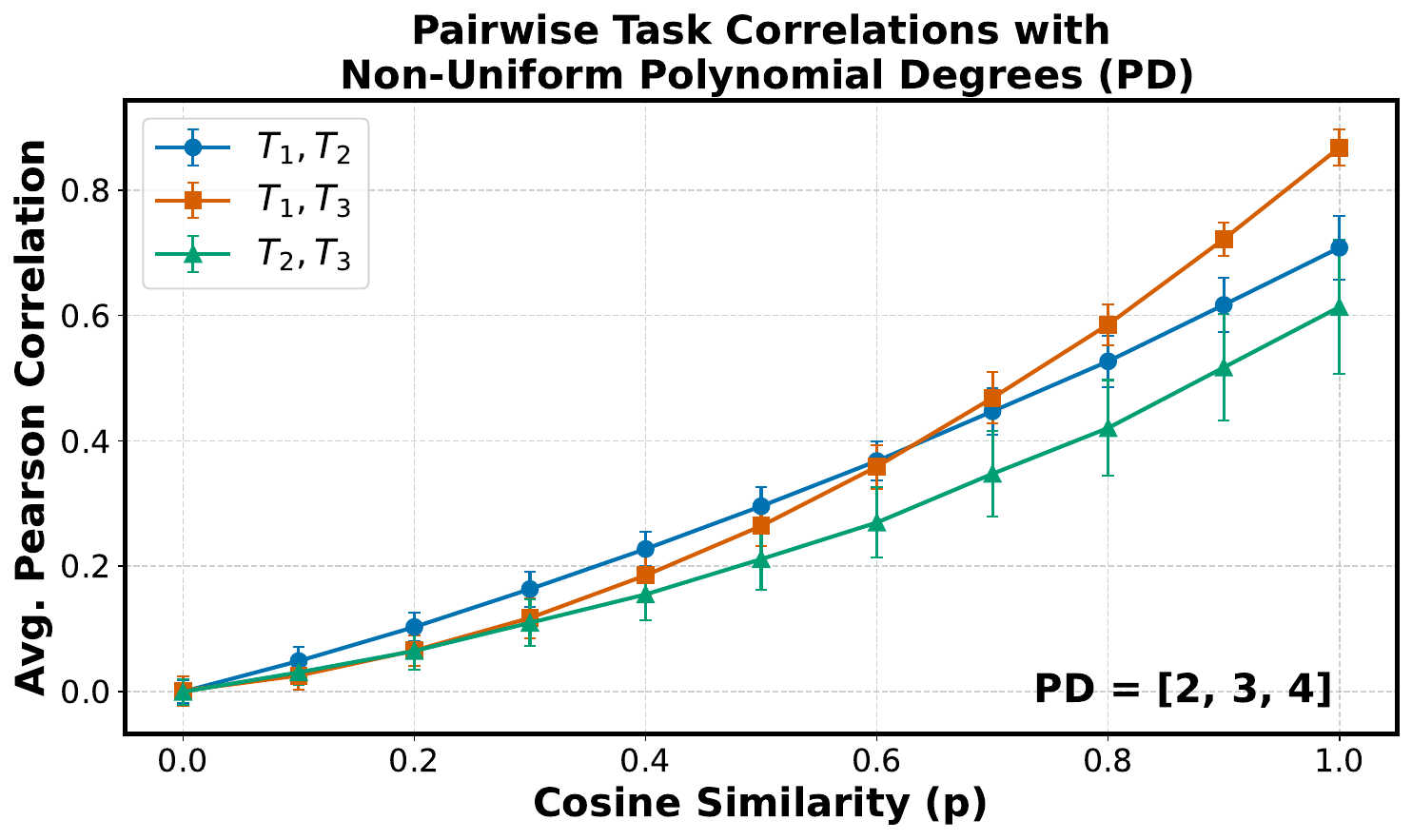}
    \caption{Average pairwise Pearson correlation for three tasks using different polynomial degrees for each task label. PD = [2, 3, 4] indicates that tasks $T_1$, $T_2$, and $T_3$ were generated using polynomial degrees 2, 3, and 4, respectively.}
    \label{fig:nonuniform}
\end{figure}

For each plot in Figures~\ref{fig:uniform} and~\ref{fig:nonuniform}, we use MultiTab-Bench to generate 100 datasets with 10k samples each and plot the mean and $2\sigma$ pairwise Pearson correlations between task labels.  In Figure~\ref{fig:uniform}, all tasks share the same polynomial degree (PD), resulting in lower label correlations as PD increases, implying a more difficult multitask learning setting. Also, the mean label correlations are the same in both plots, demonstrating that tuning task correlations is predictable regardless of the number of tasks. To achieve a more complex multitask scenario, we can simply use different PD for each task, as seen in Figure~\ref{fig:nonuniform} for a simple three-task scenario.

\begin{figure*}[htb!]
    \centering
    \includegraphics[width=0.88\linewidth]{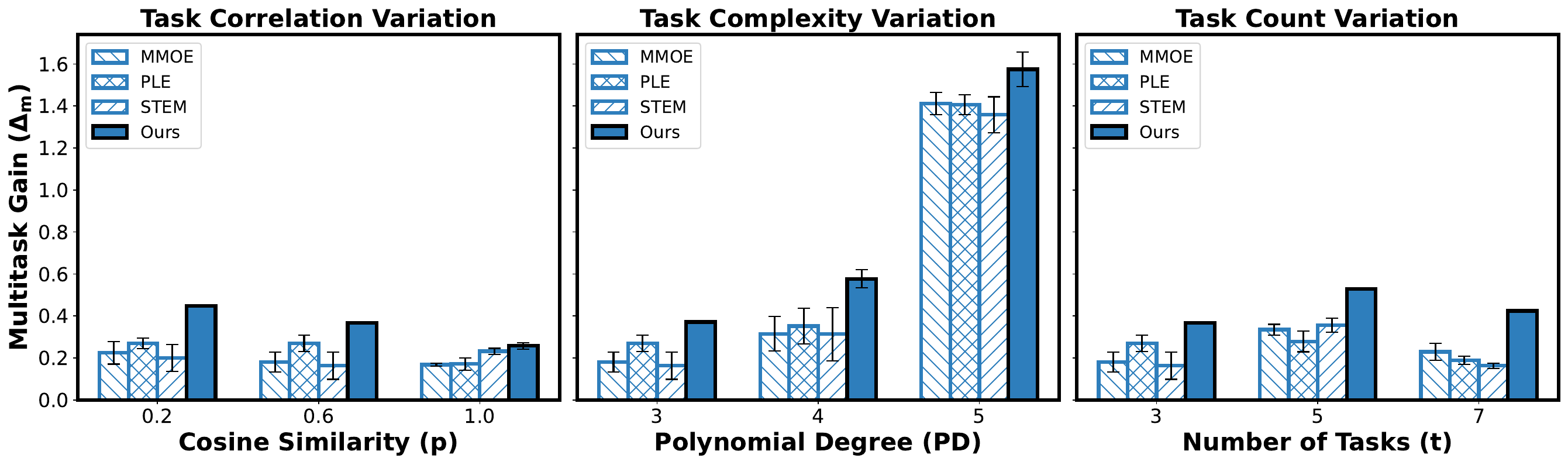}
    \caption{Average multitask gain ($\Delta_m$) comparison under controlled variations of task properties on the synthetic benchmark. Error bars represent the standard error. We vary pairwise task correlation, task complexity, and task count.
    }
    \label{fig:syn_variation}
\end{figure*}

\begin{figure}[htb!]
    \centering
   \includegraphics[width=0.95\linewidth]{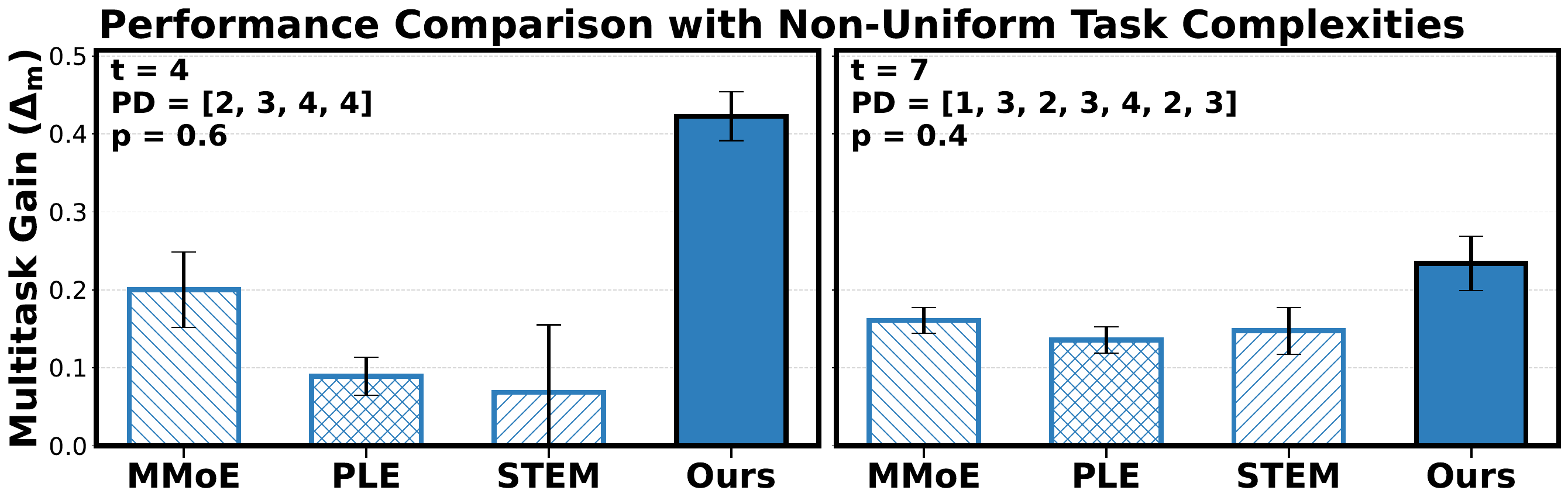}
    \caption{Average multitask gain ($\Delta_m$) comparison using non‑uniform task complexities. Error bars represent the standard error. Here, $t$ denotes the number of tasks, $\textit{PD}$ specifies the polynomial degrees used to generate labels for each task, and $p$ represents the cosine similarity controlling task correlation.}
    \label{fig:syn_complex}
\end{figure}

\section{Experimental Setup}
\label{setup}

\subsection{Public Datasets}
\s{\textbf{AliExpress}~\cite{peng2020improving} contains user behavior and transaction data, including clicks, purchases, and metadata. \s{The goal is to recommend products based on users' browsing and purchasing history.} We use the preprocessed version from~\cite{xi2021modeling}\s{, the dataset includes 12.1M training and 5.5M test samples, each with 75 features. It} which features two binary classification tasks: predicting whether a user will click on a recommended product (\textbf{Click}) and whether the click is converted into a purchase (\textbf{Conv.}).

\textbf{ACS Income}~\cite{ding2021retiring} contains survey data from the U.S. Census Bureau containing demographic and employment attributes for individual respondents. Proposed as a modern replacement for the widely used Adult Census Income dataset~\cite{adult_2}, it addresses issues such as outdated feature distributions and limited representativeness. Following prior MTL work on the Adult dataset~\cite{ma2018modeling}, we use one binary classification task predicting whether an individual’s income exceeds \$50K (\textbf{Income}) and one multi-class classification task predicting marital status (\textbf{Marital}).

\textbf{Higgs}~\cite{baldi2014searching} is a widely used single‑task benchmark for tabular learning, consisting of simulated particle collision data with kinematic features. Its main task is to distinguish signal events from background events (\textbf{Target}). In addition, a set of high‑level properties (\textbf{HLP1–HLP7})—hand‑crafted by physicists—has been shown to significantly improve single‑task performance when combined with the low‑level features. Given the scarcity of tabular benchmarks with more than two tasks, we propose a challenging multitask setting by predicting the main target alongside all HLPs.}

\textbf{AliExpress}~\cite{peng2020improving} is a large-scale recommendation dataset containing user behavior and transaction data. We use the preprocessed version from~\cite{xi2021modeling}, which includes two binary classification tasks: predicting product clicks (\textbf{Click}) and whether clicks convert to purchases (\textbf{Conv.}). \textbf{ACS Income}~\cite{ding2021retiring} is a modern replacement for the Adult Census dataset~\cite{adult_2}, featuring survey data with demographic and employment attributes. We follow prior MTL setups~\cite{ma2018modeling} with two tasks: prediction of income level being above 50k (\textbf{Income}) and marital status classification (\textbf{Marital}). \textbf{Higgs}~\cite{baldi2014searching} contains simulated particle collision data and is a popular single-task benchmark. We extend it to a multitask setting by predicting the main classification target (\textbf{Target}) alongside seven high-level properties (\textbf{HLP1–HLP7}) handcrafted by physicists, allowing evaluation under a higher task count. A summary of all dataset statistics can be found in Table~\ref{tab:dataset_summary}, which demonstrates a variety of feature counts, feature types, task counts, task types, and class balance for categorical tasks.

\begin{table}[h]
\centering
\setlength{\tabcolsep}{2pt}
\resizebox{\linewidth}{!}{%
\begin{tabular}{cccccccc}
\toprule
\multirow{2}{*}{\textbf{Dataset}} & \textbf{\# Samples (M)} &
\multicolumn{2}{c}{\textbf{\# Features}} & &
\multicolumn{2}{c}{\textbf{\# Tasks}} &
\multirow{2}{*}{\makecell{\textbf{Categorical}\\\textbf{Class Ratio (\%)}}} \\
\cline{3-4}\cline{6-7}
 & (train/val/test) & Num. & Cat. & & Num. & Cat. & \\
\midrule
Higgs & 9.90/0.55/0.55 & 17 & 4 & & 7 & 1 & \makecell{(47.01, 52.99)} \\
\midrule
AliExpress & 10.94/1.22/5.56 & 60 & 15 & & 0 & 2 & \makecell{(97.99, 2.01)\\(99.03, 0.07)} \\
\midrule
ACS Income & 1.17/0.25/0.25 & 0 & 10 & & 0 & 2 & \makecell{(60.56, 39.44)\\(54.59, 2.08, 10.74, 1.71, 30.88)} \\
\bottomrule
\end{tabular}%
}
\caption{Summary of datasets used in our experiments.}
\label{tab:dataset_summary}
\end{table}

\s{\textbf{Synthetic}~\cite{kang2011learning} tabular datasets generate samples with tunable task correlations, enabling evaluation of a multitask model's robustness to task relatedness—a key factor in multitask learning performance~\cite{duong2015low, misra2016cross}. However, existing approaches overlook other crucial factors, such as task complexity and task count. To address this, we vary (1) task correlation (\( p \)), (2) task difficulty (measured by polynomial degree, \( PD \)), and (3) task count (\( t \)) in our synthetic dataset. Each variable is isolated while keeping the others fixed to evaluate its impact on multitask performance. Details regarding the exact dataset configurations can be found in the appendix.}

\subsection{Evaluation Metrics}
For classification and regression tasks, we report the area under the ROC curve (AUC) and explained variance (EV), respectively. To evaluate overall multitask performance, we use the multitask gain ($\Delta_m$)~\cite{maninis2019attentive}, which quantifies the average improvement achieved by a multitask learning method \(m\) compared to a single task learning baseline \(b\) across all tasks \(i \in T\), as seen in equation~\ref{deltam}.
\begin{equation}
    \Delta_m = \frac{1}{T}\sum_i^T (-1)^{l_i} \left( \frac{M_{m,i} - M_{b,i}}{M_{b,i}} \right)
\label{deltam}
\end{equation}
Here, \(M_{m,i}\) and \(M_{b,i}\) denote the performance metric for method \(m\) and baseline \(b\) respectively for task \(i\). The value \(l_i\) is set to 1 if a lower metric value is desirable for task \(i\), and 0 otherwise. The resulting \(\Delta_m\) is interpreted as a percentage of the average improvement over the baseline \(b\). Additional metrics and variance analysis can be found in the appendix.

\subsection{Baselines}
For the single-task baseline (\textbf{STL}), we follow the standard practice in MTL by using independent multi-layer perceptrons (MLPs)~\cite{goodfellow2016deep} for each task. The multitask baseline (\textbf{MTL}) follows the ``shared-bottom" architecture~\cite{caruana1997multitask}, where a portion of neurons is shared across tasks. We also compare against state-of-the-art multitask models used in large-scale recommendation systems: \textbf{MMoE}~\cite{ma2018modeling}, \textbf{PLE}~\cite{tang2020progressive}, and \textbf{STEM}~\cite{su2024stem}, which are all MLP-based.\s{ These models, built upon MLPs for their training efficiency, have been widely deployed in real-world applications.} As the first multitask transformer for tabular data, there is no direct multitask baseline, so we will also evaluate single-task tabular transformers, including \textbf{TabT}~\cite{huang2020tabtransformer}, \textbf{FT-T}~\cite{gorishniy2021revisiting}, and \textbf{SAINT}~\cite{somepalli2021saint}. Lastly, we include \textbf{XGBoost}~\cite{chen2016xgboost} due to its strong performance on tabular data.

\subsection{Implementation Details}
We implement all baseline models from scratch, following open‑source implementations closely when available to ensure a fair and unified experimental setting. It has been shown that neural networks typically require extensive tuning for each tabular dataset~\cite{mcelfresh2023neural}, so we perform a thorough hyperparameter sweep for every model using grid search with \texttt{WANDB}~\cite{wandb}. Details of the sweep configurations and the selected parameters are provided in the appendix. To make the evaluation fair, we keep model capacities comparable across methods, although we note that simply increasing capacity does not always lead to better performance. For consistency, we transform the inputs to embeddings for all baselines, as prior work has shown this to improve tabular performance~\cite{somepalli2021saint,su2024stem}. All models are trained with the Adam optimizer~\cite{kingma2014adam} and weight decay. Results are aggregated over five random seeds, with the same five seeds used for every model and dataset. All experiments are run on a single RTX A5500 GPU. Our full \texttt{PyTorch Lightning} implementation is included in the supplementary material.

\begin{table}[htb!]
\centering
\setlength{\tabcolsep}{3pt}
\resizebox{0.9\linewidth}{!}{%
\begin{tabular}{c c c c c c c c}
\toprule
\multicolumn{2}{c}{MultiTab-Net} & & \textit{AliExpress} & & \textit{ACS Income} & & \textit{Higgs} \\
\cline{1-2}\cline{4-4}\cline{6-6}\cline{8-8}\\[-2ex]
\# Task Tokens & Attn. Mask & & $\Delta_m \uparrow$ & & $\Delta_m \uparrow$ & & $\Delta_m \uparrow$ \\
\midrule
\multirow{2}{*}{Single} 
  & No Masking  & & 0.2669 & & 0.0893 & & -6.3491 \\
  & F$\not\to$T & & 0.1301 & & 0.0837 & & -12.5587 \\
\midrule
\multirow{4}{*}{Multiple} 
  & No Masking               & & 0.2579 & & 0.0783 & & 1.1182 \\
  & F$\not\to$T \& T$\not\to$T & & 0.2975 & & 0.1007 & & 1.0197 \\
  & F$\not\to$T              & & 0.3698 & & 0.0951 & & 0.9626 \\
  & T$\not\to$T              & & \textbf{0.5512} & & \textbf{0.1064} & & \textbf{1.2337} \\
\bottomrule
\end{tabular}}
\caption{Ablation of multitask attention (Attn.) masking methods for MultiTab-Net using a single shared task token and multiple task tokens (i.e., one for each task). When using a single task token, only F$\not\to$T is possible. Boldface indicates the best value per column.}
\label{tab:ablation}
\end{table}

\begin{table*}[htb!]
\centering
\setlength{\tabcolsep}{3pt}
\resizebox{1.0\textwidth}{!}{%
\begin{tabular}{@{}ccccccrcccccrcccccccccccccccccr@{}}
\toprule
 &  & \multicolumn{5}{c}{\textit{AliExpress}} &  & \multicolumn{5}{c}{\textit{ACS Income}} &  & \multicolumn{15}{c}{\textit{Higgs}} &  &  \\ \cline{3-7}\cline{9-13}\cline{15-31}\\[-2ex]
\multirow{2}{*}{Models} &  & \textbf{Click} &  & \textbf{Conv}. &  & \multirow{2}{*}{$\Delta_m$ ↑} &  & \textbf{Income} &  & \textbf{Marital} &  & \multirow{2}{*}{$\Delta_m$ ↑} &  & \textbf{Target} &  & \textbf{HLP1} &  & \textbf{HLP2} &  & \textbf{HLP3} &  & \textbf{HLP4} &  & \textbf{HLP5} &  & \textbf{HLP6} &  & \textbf{HLP7} &  & \multirow{2}{*}{$\Delta_m$ ↑} \\
 &  & AUC ↑ &  & AUC ↑ &  &  &  & AUC ↑ &  & AUC ↑ &  &  &  & AUC ↑ &  & EV ↑ &  & EV ↑ &  & EV ↑ &  & EV ↑ &  & EV ↑ &  & EV ↑ &  & EV ↑ &  &  \\
\cline{1-1}\cline{3-7}\cline{9-13}\cline{15-31}\\[-2ex]
STL &  & 72.07 &  & 85.67 &  & 0.0000 &  & 90.19 &  & 88.54 &  & 0.0000 &  & 84.90 &  & 94.39 &  & 32.42 &  & 38.79 &  & 60.43 &  & 99.19 &  & 68.16 &  & 62.83 &  & 0.0000 \\
MTL &  & 72.30 &  & 85.59 &  & 0.1129 &  & 90.28 &  & 88.56 &  & 0.0612 &  & 84.13 &  & 93.71 &  & 32.66 &  & 38.60 &  & 59.34 &  & 96.67 &  & 68.39 &  & 62.93 &  & -0.6531 \\
MMoE &  & 72.28 &  & 85.57 &  & 0.0873 &  & 90.29 &  & 88.60 &  & 0.0893 &  & 85.03 &  & 93.98 &  & 32.34 &  & 38.48 &  & 59.95 &  & 97.91 &  & 68.39 &  & 62.99 &  & -0.3525 \\
PLE &  & 72.42 &  & 85.73 &  & 0.2778 &  & 90.30 &  & 88.59 &  & 0.0892 &  & 85.31 &  & 94.31 &  & 32.37 &  & 38.84 &  & 60.27 &  & 98.20 &  & 68.40 &  & 63.01 &  & -0.0314 \\
TabT &  & 72.15 &  & 85.91 &  & 0.1956 &  & 90.16 &  & 88.51 &  & -0.0336 &  & 72.61 &  & 78.90 &  & 26.10 &  & 28.66 &  & 37.54 &  & 57.23 &  & 54.90 &  & 48.88 &  & -24.7917 \\
FT-T &  & 72.28 &  & 85.87 &  & 0.2624 &  & 90.13 &  & 88.46 &  & -0.0784 &  & 78.38 &  & 94.42 &  & 31.59 &  & 36.52 &  & 54.05 &  & \textbf{99.87} &  & 66.85 &  & 63.05 &  & -3.4380 \\
SAINT &  & 72.21 &  & 85.70 &  & 0.1146 &  & \textbf{90.31} &  & 88.59 &  & 0.0948 &  & 85.54 &  & \textbf{94.44} &  & 31.14 &  & 37.14 &  & 57.96 &  & 98.14 &  & 67.66 &  & 62.87 &  & -1.6514 \\
XGB &  & 72.15 &  & \textbf{86.02} &  & 0.2598 &  & 89.91 &  & 88.47 &  & -0.1948 &  & 73.45 &  & 79.36 &  & 26.51 &  & 30.79 &  & 43.80 &  & 63.50 &  & 62.73 &  & 57.37 &  & -18.5526 \\
STEM &  & 72.24 &  & 85.77 &  & 0.1763 &  & 90.28 &  & 88.58 &  & 0.0725 &  & 85.32 &  & 94.29 &  & 32.59 &  & 38.78 &  & 60.36 &  & 98.35 &  & 68.36 &  & 62.98 &  & 0.0571 \\
MultiTab-Net &  & \textbf{72.57} &  & \textbf{86.02} &  & \textbf{0.5512} &  & 90.28 &  & \textbf{88.64} &  & \textbf{0.1064} &  & \textbf{85.99} &  & 93.99 &  & \textbf{33.63} &  & \textbf{39.82} &  & \textbf{61.36} &  & 97.96 &  & \textbf{68.93} &  & \textbf{63.58} &  & \textbf{1.2337} \\ \bottomrule
\end{tabular}}
\caption{Test set results for all models across all public datasets. Boldface indicates the best value per column.}
\label{tab:public_results}
\end{table*}

\section{Results}
\label{results}
\s{In this section, we will discuss the results from our ablation studies, our comparison against state-of-the-art models on public data, and our experiments on our proposed generalized multitask synthetic dataset.}

\subsection{Ablation Study}
To assess the effectiveness of our architecture and candidate masking strategies, we conduct an ablation across all three public datasets, varying the number of task tokens (single shared vs. multiple) and the candidate attention masks. As shown in Table~\ref{tab:ablation}, two key trends emerge. First, the advantage of using multiple task tokens increases with the number of tasks. While the difference is minimal on AliExpress and ACS Income (each with two tasks), the multi-token setup yields clear gains on the eight-task Higgs dataset. This suggests that a single shared token cannot adequately capture task-specific information in more complex multitask settings. Second, the choice of mask strongly affects performance. Notably, F$\not\to$T offers inconsistent benefits, implying that features should retain access to task context. In contrast, T$\not\to$T consistently performs the best, showing that preventing task tokens from attending to each other helps reduce task interference. Accordingly, all subsequent experiments use the multi-token configuration with T$\not\to$T masking.

\subsection{Experiments with MultiTab-Bench}
Our synthetic tabular data generator enables controlled variation of key multitask factors, allowing us to evaluate the effects of task correlation ($p$), task complexity (measured by polynomial degree, $PD$), and task count ($t$). Since it is not feasible to test every possible configuration, we isolate the effect of each factor by varying one at a time while keeping the others fixed. As shown in Figure~\ref{fig:syn_variation}, MultiTab-Net (ours) consistently achieves higher multitask gain than existing baselines (MMoE, PLE, STEM) across all variations, demonstrating robustness to changes in task relatedness, scalability with increasing task count, and improved handling of task complexity. We further evaluate two settings where tasks have markedly different levels of task complexity and observe in Figure~\ref{fig:syn_complex} that MultiTab-Net (ours) again outperforms all baselines by a large margin. These results highlight that MultiTab-Net not only handles balanced multitask scenarios well but also excels when tasks vary widely in difficulty, which is common in real‑world applications. Full dataset configurations and experimental details are provided in the appendix.

\subsection{Comparison to State-of-the-Art}
The results in Table~\ref{tab:public_results} show that MultiTab-Net consistently achieves the highest multitask gain, $\Delta_m$, across all datasets. Notably, MultiTab-Net consistently outperforms STEM, the most recent multitask model for tabular data, highlighting the strength of a transformer architecture for multitask learning on large-scale tabular data. The comparison with SAINT, which MultiTab-Net extends into a multitask framework, further shows the value of our design choices. Despite their architectural similarities, SAINT shows limited or even negative multitask gain, whereas MultiTab-Net achieves clear improvements. On Higgs, MultiTab-Net again delivers the strongest results, achieving a substantial $\Delta_m$. XGBoost’s poor performance aligns with findings in the original Higgs paper, where tree‑based models struggled to capture the complex targets. TabT also performs poorly on Higgs because the dataset contains only four categorical features, leaving most of the architecture reduced to a shallow MLP—a limitation documented in its original work. Overall, these results demonstrate that MultiTab-Net’s combination of multi‑token design and T$\not\to$T masking leads to consistently better generalization and scalability across diverse tabular benchmarks.

\subsection{Computational Efficiency Analysis}

In Table~\ref{tab:efficiency}, we evaluate the computational requirements for MultiTab-Net against the other multitask tabular models (MMoE, PLE, STEM), and MultiTab-Net's closest single-task counterpart (SAINT). When comparing to other MTL methods, it is clear that we have a substantial advantage in computational efficiency in ACS Income and Higgs, however, MultiTab-Net is slightly more expensive in AliExpress due to a high number of features. When comparing to SAINT, which shares many architectural similarities to MultiTab-Net, we achieve an efficiency multiplier roughly equal to the number of tasks (i.e., roughly 2x fewer cost on AliExpress and ACS Income, and 8x on Higgs), highlighting the benefits of multitask learning.

\begin{table}[htb!]
\centering
\setlength{\tabcolsep}{2pt}
\resizebox{1.0\linewidth}{!}{%
\begin{tabular}{@{}c c c c c c c c@{}}
\toprule
\multirow{2}{*}{Model} &  & \textit{AliExpress} &  & \textit{ACS Income} &  & \textit{Higgs} &  \\
\cline{3-3} \cline{5-5} \cline{7-7}\\[-2ex]
&  & Params/FLOPs (M) &  & Params/FLOPs (M) &  & Params/FLOPs (M) &  \\
\midrule
SAINT     &  & 3.62 / 9.70  &  & 0.49 / 1.35  &  & 5.50 / 15.02  &  \\
MMOE      &  & \textbf{1.53 / 3.07}  &  & 0.64 / 1.25  &  & 1.05 / 2.11  &  \\
PLE       &  & 1.55 / 3.11  &  & 0.65 / 1.28  &  & 1.22 / 2.46  &  \\
STEM      &  & 1.55 / 3.11  &  & 0.69 / 1.29  &  & 1.25 / 2.51  &  \\
MultiTab-Net  &  & 1.80 / 4.85  &  & \textbf{0.28 / 0.77}  &  & \textbf{0.70 / 1.90}  &  \\
\bottomrule
\end{tabular}}
\caption{Number of parameters (Params) and floating point operations (FLOPs) in millions (M). Lower values are desirable. Boldface indicates the best result in each column.}
\label{tab:efficiency}
\end{table}

\section{Conclusion}
\label{conclusion}
We introduced MultiTab-Net, the first transformer-based architecture for multitask learning on tabular data. Its innovative multi-token approach with the multitask masked-attention effectively limits task interference while capturing complex feature interactions. MultiTab-Net consistently outperformed existing MTL baselines across diverse datasets and scaled well with increasing task count and complexity. We also proposed MultiTab-Bench, a synthetic multitask dataset generator for systematic evaluation of key MTL factors, including task correlation, complexity, and count. Together, these contributions advance the state of multitask learning in tabular domains and establish new benchmarks for future research.

\bibliography{aaai2026}


\appendix
\onecolumn
\section{Appendix}

\subsection{More Details on Related Works}
\subsubsection{Tabular Deep Learning}

Deep learning has achieved remarkable success with high-dimensional and homogeneous domains that use image and text data ~\cite{krizhevskyImageNetClassificationDeep2012, devlin2018bert}, but remains less prominent in tabular data, where gradient-boosted decision trees (GBDTs)~\cite{chen2016xgboost} excel at handling mixed-type heterogeneous features and irregular feature distributions. Despite this, recent works has demonstrated that deep learning approaches achieve competitive or even superior performance on tabular benchmarks~\cite{mcelfresh2023neural, ericksonTabArenaLivingBenchmark2025a}.

\s{\paragraph{MLP-based models.}}
Multi-Layer Perceptrons (MLPs) serve as a fundamental deep learning baseline for tabular data. While simple, they achieve competitive results when paired with techniques such as advanced regularization, extensive tuning  \cite{kadra2021well} and feature-wise embeddings~\cite{gorishniy2022embeddings}. 

\s{\paragraph{Transformer-based models.}}
Transformer-based models have recently emerged as strong alternatives in tabular data, inspired by their success in the image and text domains~\cite{devlin2018bert, dosovitskiy2020image}. TabTransformer (TabT)~\cite{huang2020tabtransformer} embeds categorical features with self-attention layers before passing them to an MLP for prediction. FT-Transformer (FT-T)~\cite{gorishniy2021revisiting} embeds both discrete and continuous features into tokens before jointly processing them with transformer layers. SAINT~\cite{somepalli2021saint} extends this framework by introducing inter-sample attention layers, enabling information sharing across rows within a batch, which has been shown to deliver consistent performance improvements on tabular datasets. Recent advances such as TabPFN ~\cite{hollmann2022tabpfn, hollmann2025accurate} and TabICL ~\cite{qu2025tabicl} leverage the transformer architecture and introduce tabular foundation models that use in‑context learning by conditioning on labeled examples at inference. However, they were designed for datasets with up to 10k and 60k samples respectively, making their in‑context framework inapplicable to our large‑scale setting.





\subsubsection{Multitask Learning on Tabular Data}

\s{Multitask learning (MTL) remains largely underexplored in tabular data, with most research focused on recommendation systems. A key advancement in this space is MMoE~\cite{ma2018modeling}, which introduced a mixture of experts (MoE) framework using an MLP backbone. In this approach, shared neurons are partitioned into experts, whose outputs are dynamically weighted by gates before being processed by task-specific towers. PLE~\cite{tang2020progressive} extended this by segmenting experts into shared and task-specific groups. Task-specific gates accessed only their corresponding experts and the shared experts, while shared gates aggregated all experts. This design mitigated cross-task interference, commonly referred to as the ``seesaw phenomenon," where performance improvements in one task degrade another.\s{ PLE also introduced stacked expert layers to enhance model flexibility and expressiveness.} The most recent advancement, STEM~\cite{su2024stem}, builds on PLE by incorporating feature-wise embeddings, inspired by prior work in single-task tabular MLPs~\cite{gorishniy2022embeddings}. Unlike PLE, STEM applies gating mechanisms that aggregate outputs from all experts in the forward pass while introducing stop-gradient constraints between gates and experts corresponding to different tasks in the backward pass. This prevents unintended updates to task-specific embeddings, further reducing interference. This principle directly influenced our multitask masked attention mechanism in MultiTab-Net. To the best of our knowledge, no prior work has explored transformer-based MTL for tabular data. MultiTab-Net is the first such model, extending transformers to MTL in this domain and marking a significant step forward.}

Multitask learning (MTL) in tabular data remains underexplored beyond recommendation systems. MMoE~\cite{ma2018modeling} introduced a mixture-of-experts (MoE) approach using MLPs, where expert outputs are gated and fed into task-specific towers. PLE~\cite{tang2020progressive} improved this by adding shared and task-specific experts, with gating schemes designed to reduce cross-task interference, also known as the “seesaw phenomenon.” STEM~\cite{su2024stem} further refined this by using feature-wise embeddings and gating all experts in the forward pass while applying stop-gradient constraints in the backward pass to isolate task-specific updates. This inspired the multitask masked attention in MultiTab-Net. To our knowledge, MultiTab-Net is the first transformer-based MTL model for tabular data, representing a novel architectural direction in this space.

\subsection{Inter-Sample Attention}
The inter-sample attention module aims to model dependencies across different samples in a batch. Unlike inter-feature attention, this module flattens the feature and embedding dimensions for each sample, including task tokens, before computing attention. This allows the model to capture global interactions between samples. For a batch of $n$ samples of $x \in \mathbb{R}^{(d + t) \times e}$, we flatten each sample along the feature and embedding dimensions to obtain $X \in \mathbb{R}^{n \times ((d + t) e)}$. We compute the attention scores and output the same way as in inter-feature attention by substituting $X$ in place of $x$ in Equation~\ref{eq:qkv}. Consequently, we use a different set of learnable projection matrices, $W_{Q_i}, W_{K_i}, W_{V_i} \in \mathbb{R}^{((d + t) e) \times d_k}$, than the ones used in inter-sample attention. Also for inter-sample attention, $A_i \in \mathbb{R}^{n \times n}$ captures the attention scores among samples in the batch. In the inter-sample multi-head setting, the input is split across $h$ heads, such that $d_k = \frac{((d + t) e)}{h}$ for each head, and the outputs from all heads are concatenated and multiplied with a different output project layer $W_O \in \mathbb{R}^{((d + t) e) \times ((d + t) e)}$ to form the final representation.

\subsection{Overview of MMoE's Synthetic Data Generation}
The synthetic dataset simulates two regression tasks, where the task relatedness is quantified using the Pearson correlation of their labels. 

\begin{enumerate}
\item Given feature dimension \(d\), sample two orthonormal vectors,  \(u_1, u_2 \in \mathbb{R}^{d}\), i.e. \(u_1^T u_2 =0\), \( \| u_1 \| = 1 \), \( \| u_2 \| = 1 \).
\item Define task-specific weight vectors with a given correlation score \( -1 \leq p \leq 1 \):
\[ w_1 = c u_1, \quad w_2 = c (p u_1 + \sqrt{1 - p^2} u_2). \]

\item Sample input features \( x \sim \mathcal{N}(0, I) \) from a standard normal distribution.

\item Generate regression task labels using sinusoidal transformations:
\[ y_1 = w_1^T x + \sum_{i=1}^{m} \sin(\alpha_i w_1^T x + \beta_i) + \epsilon_1, \]
\[ y_2 = w_2^T x + \sum_{i=1}^{m} \sin(\alpha_i w_2^T x + \beta_i) + \epsilon_2, \]
where \( \alpha_i, \beta_i \) control the sinusoidal functions, and noise terms \( \epsilon_1, \epsilon_2 \sim \mathcal{N}(0, 0.01) \) introduce uncertainty to the labels.
\item Repeat 3 and 4 \(n\) times to obtain \(n\) total samples.
\end{enumerate}

By construction, the Pearson correlation of the labels,\(y_1\), \(y_2\), is positively correlated with the cosine similarity of the weight vectors \(w_1\), \(w_2\). This relationship was empirically validated in~\cite{ma2018modeling}, demonstrating that as the cosine similarity of the weight vectors increases, the corresponding label correlation increases accordingly. 

\subsection{Proof of Cosine Similarity}
Recall from Section~4.1 that the task-specific weight matrix, \(\mathbf{W} \in \mathbb{R}^{t \times d}\), is defined by

\begin{equation}
    \mathbf{W} = \mathbf{Q} \boldsymbol{\Lambda}^{1/2} \mathbf{U}^T,
\end{equation}

where \(\mathbf{Q}\) is an orthogonal matrix of eigenvectors of \(\mathbf{P}\), i.e. \( \mathbf{P} = \mathbf{Q} \boldsymbol{\Lambda} \mathbf{Q}^T \) and \(\boldsymbol{\Lambda}^{1/2}\) is the diagonal matrix of square roots of eigenvalues, i.e.\ \(\boldsymbol{\Lambda}^{1/2} = \mathrm{diag}(\sqrt{\lambda_1},\dots,\sqrt{\lambda_t})\),
and \(\mathbf{U} \in \mathbb{R}^{d \times t}\) is a matrix with orthonormal columns such that \(\mathbf{u}_k^T\mathbf{u}_\ell = \delta_{k\ell}\). \(\delta_{k\ell}\) is defined as follows,
\[
\delta_{k\ell} =
\begin{cases}
1, & \text{if } k = \ell, \\
0, & \text{if } k \neq \ell.
\end{cases}
\]

Expanding \(\mathbf{w}_i\):
\begin{equation}
    \mathbf{w}_i = \sum_{k=1}^t Q_{ik} \sqrt{\lambda_k} \mathbf{u}_k.
\end{equation}
Expanding this product and using the orthonormality of \(\{\mathbf{u}_k\}\), we obtain
\begin{equation}
\mathbf{w}_i^T \mathbf{w}_j
=\;
\sum_{k=1}^t \sum_{\ell=1}^t 
Q_{ik}\,Q_{j\ell}\,\sqrt{\lambda_k}\,\sqrt{\lambda_\ell}\,\mathbf{u}_k^T \mathbf{u}_\ell
\\
=\;
\sum_{k=1}^t \sum_{\ell=1}^t
Q_{ik}\,Q_{j\ell}\,\sqrt{\lambda_k}\,\sqrt{\lambda_\ell}\,\delta_{k\ell}
\\
=\;
\sum_{k=1}^t
Q_{ik}\,Q_{jk}\,\lambda_k.
\end{equation}
Using the eigendecomposition \(\mathbf{P} = \mathbf{Q} \Lambda \mathbf{Q}^T\), we obtain:
\begin{equation}
    \mathbf{w}_i^T \mathbf{w}_j = P_{ij}.
\end{equation}
Since \(\|\mathbf{w}_i\| = 1\), the cosine similarity is:
\begin{equation}
    \cos(\mathbf{w}_i, \mathbf{w}_j) = \frac{\mathbf{w}_i^T \mathbf{w}_j}{\|\mathbf{w}_i\| \|\mathbf{w}_j\|} = P_{ij}.
\end{equation}

\subsection{Implementation Details}
We reimplemented all baseline models from scratch in a unified PyTorch Lightning codebase, following open-source reference implementations and official repositories wherever available. This approach allowed us to maintain consistency across preprocessing, data handling, and model evaluation, while also applying a uniform logging and tuning framework. For the AliExpress dataset, we used the preprocessed train/test splits from the MTLRecLib repository. Since no official validation set was provided, we created one by stratifying 10\% of the training set. For ACS Income, we randomly shuffled the data and split it into 70\% training, 15\% validation, and 15\% testing. For Higgs, we followed prior work, using the last 5\% of the data for testing, the first 5\% for validation, and the remaining 90\% for training. All experiments used early stopping, with a patience of 3 epochs for AliExpress and 5 epochs for ACS Income and Higgs. Given the multitask nature of our experiments, we avoided using a metric from a single task to monitor convergence. Instead, we tracked the average AUROC for classification tasks and average Explained Variance (EV) for regression tasks, leveraging the fact that both metrics are bounded between 0 and 1. This approach allowed us to define a consistent and fair early stopping criterion across all datasets. The best-performing model on the validation set was saved and later evaluated on the held-out test set, and all reported metrics in this paper reflect this final evaluation. To facilitate fair comparison between architecture types (MLPs, transformers, and expert-based models), we matched embedding sizes and model capacities across baselines. We used an embedding size of 16 for ACS Income and Higgs, and 8 for AliExpress (due to the higher number of input features). All models used the Adam optimizer with a weight decay of 1e-5 and a batch size of 2048. Transformer-based models used a hidden size of 256, while MLP-based models used a two-layer configuration with sizes [256, 128], which we found led to comparable model capacities and depths. For models using expert-based architectures (MMoE, PLE, STEM), we followed prior work and used 8 experts. Transformer-based models (including MultiTab-Net, SAINT, FT-T and TabT) used 4 attention heads. All deep learning models were tuned using a grid search with values \{0.0001, 0.001, 0.01, 0.1\} for learning rate and \{0.0, 0.1, 0.2, 0.3\} for dropout. For transformer-based models, we separately tuned attention and feedforward dropout rates. For MultiTab-Net and SAINT, we introduced an additional hyperparameter: the use of rotary positional embeddings (RoPE)~\cite{su2024roformer} within the inter-sample attention layers. Inspired by TabICL’s~\cite{qu2025tabicl} findings, we found that using RoPE in these layers helped stabilize training and avoid representation collapse, particularly for AliExpress, Higgs, and settings explored in MultiTab-Bench. This optional component significantly improved MultiTab-Net's performance in certain settings. For XGBoost, we used the multioutput variant whenever the dataset contained tasks of the same type—such as the two binary classification tasks in AliExpress or the seven regression tasks in Higgs. In mixed-task settings like ACS Income, we trained separate vanilla XGBoost models for each task. We performed a grid search over maximum depth \{3, 4, 5, 6, 7, 8, 9, 10\}, learning rate \{0.01, 0.02, 0.05, 0.1, 0.2\} (where higher rates often yielded better results), and number of boosting rounds \{25, 50, 100, 200, 300\}. The training objective was set to binary cross-entropy for binary classification, categorical cross-entropy for multiclass classification, and RMSE for regression tasks. Full sweep configurations and final selected parameters per model and dataset are available in the shared codebase.

\subsection{Dataset Details}
\textbf{AliExpress}~\cite{peng2020improving} contains user behavior and transaction data, including clicks, purchases, and metadata. \s{The goal is to recommend products based on users' browsing and purchasing history.} We use the preprocessed version from~\cite{xi2021modeling}\s{, the dataset includes 12.1M training and 5.5M test samples, each with 75 features. It} which features two binary classification tasks: predicting whether a user will click on a recommended product (\textbf{Click}) and whether the click is converted into a purchase (\textbf{Conv.}).

\textbf{ACS Income}~\cite{ding2021retiring} contains survey data from the U.S. Census Bureau containing demographic and employment attributes for individual respondents. Proposed as a modern replacement for the widely used Adult Census Income dataset~\cite{adult_2}, it addresses issues such as outdated feature distributions and limited representativeness. Following prior MTL work on the Adult dataset~\cite{ma2018modeling}, we use one binary classification task predicting whether an individual’s income exceeds \$50K (\textbf{Income}) and one multi-class classification task predicting marital status (\textbf{Marital}).

\textbf{Higgs}~\cite{baldi2014searching} is a widely used single‑task benchmark for tabular learning, consisting of simulated particle collision data with kinematic features. Its main task is to distinguish signal events from background events (\textbf{Target}). In addition, a set of high‑level properties (\textbf{HLP1–HLP7})—hand‑crafted by physicists—has been shown to significantly improve single‑task performance when combined with the low‑level features. Given the scarcity of tabular benchmarks with more than two tasks, we propose a challenging multitask setting by predicting the main target alongside all HLPs.

\subsection{Comparison to Prior Experimental Settings in Tabular MTL}

Recent work on multitask learning (MTL) for tabular data has adopted varying experimental setups, particularly in terms of dataset accessibility, task structure, and evaluation scope. Our experimental design aligns more closely with the framework introduced by \textbf{MMoE} and followed by \textbf{PLE}, rather than the more recent setting used by \textbf{STEM}, for the following reasons.

MMoE evaluated their model on three datasets: (1) a modified version of the Census-Income dataset, where a second task (Marital Status) was added to enable MTL; (2) a proprietary recommendation dataset from Google; and (3) a synthetic benchmark for analyzing task relationships, limited to two-task settings. PLE followed a similar protocol, using the same synthetic dataset and Census-Income benchmark, along with two recommendation datasets: the public Ali-CCP dataset and a proprietary Tencent dataset.

STEM diverged from this pattern and instead evaluated on three large-scale public recommendation datasets—TikTok, QK-Video, and KuaiRand1K—as well as a private dataset from Tencent. While this reflects industrial-scale applications, it raises concerns for general-purpose MTL research. First, only the dataset link and preprocessing code for TikTok is accessible via their codebase, and even then, acquiring the data is still difficult for researchers outside of China. For the remaining datasets, neither preprocessing code nor download instructions are provided, making their setup hard to reproduce. Second, these datasets contain extremely high-cardinality categorical features, resulting in massive embedding tables that dominate model parameter count (over 99\%). This makes the contribution of downstream MTL architectures relatively minor. Third, STEM uses both shared and task-specific embeddings, meaning that multiple large embedding tables must be maintained—one for each task plus a shared table—dramatically increasing model capacity even in low task-count settings. This results in an imbalance that favors their architectural design and complicates fair comparison.

In our work, we follow the MMoE and PLE evaluation pattern, while modernizing and extending it in several ways. First, we use an updated version of the Census-Income dataset with the same MTL structure (Income and Marital Status). Second, we include the large-scale AliExpress dataset, which is well-documented and available preprocessed through the MTLRecLib repository. This dataset is more accessible than those used in STEM and better suited to reproducible experimentation. Third, we extend MMoE’s synthetic benchmark by introducing a new synthetic generator that allows fine-grained control over task count, pairwise task correlation, and relative task complexity—enabling broader and more rigorous analysis of multitask dynamics. Finally, we introduce a new multitask benchmark using the popular Higgs dataset (easily accessible via OpenML or the UCI repository), where we supplement the original binary classification task with handcrafted high-level physics features as auxiliary targets. This produces a realistic multitask benchmark that addresses the scarcity of publicly available tabular MTL datasets with more than two tasks.

In summary, our evaluation design emphasizes reproducibility, accessibility, and flexibility. While STEM targets industrial-scale recommender systems, our setting better supports general-purpose multitask learning on tabular data by ensuring balanced model capacity, transparent preprocessing, and broad applicability.

\subsection{Statistical Significance Analysis}
In this study, we benchmark MultiTab-Net across three distinct multitask tabular datasets: AliExpress, which focuses on click-through and conversion prediction; ACS Income, a two-task demographic prediction dataset including income (binary classification) and marital status (multiclass classification); and Higgs, a complex dataset with one classification task and seven regression tasks based on high-level physics features. For each dataset, we report the mean ($\mu$) and standard deviation ($\sigma$) of the performance metrics across five random seeds. These results, shown in Tables~\ref{tab:aliexpress-results}, \ref{tab:inc_extra}, and \ref{tab:higgs_ev_scaled}, are extended versions of the main paper’s results tables and now include variance information to support deeper statistical analysis.

\begin{table}[h!]
\setlength{\tabcolsep}{4pt}
\centering
\begin{tabular}{@{}ccccccccccc@{}}
\toprule
\multirow{3}{*}{Model} &  & \multicolumn{2}{c}{Click AUC} &  & \multicolumn{2}{c}{Conv. AUC} &  & \multicolumn{2}{c}{$\Delta_m$} \\
\cline{3-4}\cline{6-7}\cline{9-10}\\[-2ex]
 &  & $\mu$ & $\sigma$ &  & $\mu$ & $\sigma$ &  & $\mu$ & $\sigma$ \\ \midrule
STL   &  & 72.07 & 0.32 &  & 85.67 & \textbf{0.19} &  & 0.0000 & 0.2413 \\
MTL   &  & 72.30 & 0.21 &  & 85.59 & 0.64         &  & 0.1129 & 0.4284 \\
MMoE  &  & 72.28 & 0.42 &  & 85.57 & 0.43         &  & 0.0873 & 0.3268 \\
PLE   &  & 72.42 & 0.54 &  & 85.73 & 0.46         &  & 0.2778 & 0.5328 \\
TabT  &  & 72.15 & 0.19 &  & 85.91 & 0.30         &  & 0.1956 & 0.2876 \\
FT-T   &  & 72.28 & 0.31 &  & 85.87 & 0.24         &  & 0.2624 & \textbf{0.0836} \\
SAINT &  & 72.21 & 0.27 &  & 85.70 & 0.39         &  & 0.1146 & 0.3610 \\
STEM  &  & 72.24 & 0.23 &  & 85.77 & 0.47         &  & 0.1763 & 0.3507 \\
MultiTab-Net   &  & \textbf{72.57} & \textbf{0.12} &  & \textbf{86.02} & 0.54 &  & \textbf{0.5512} & 0.2459 \\ \bottomrule
\end{tabular}
\caption{Results on AliExpress, including mean ($\mu$) and standard deviation ($\sigma$) for binary AUC on Click and Conversion tasks across 5 seeds.}
\label{tab:aliexpress-results}
\end{table}

\begin{table}[h!]
\setlength{\tabcolsep}{4pt}
\centering
\begin{tabular}{@{}ccccccccccc@{}}
\toprule
\multirow{3}{*}{Model} &  & \multicolumn{2}{c}{Income AUC} &  & \multicolumn{2}{c}{Marital AUC} &  & \multicolumn{2}{c}{$\Delta_m$} \\
\cline{3-4}\cline{6-7}\cline{9-10}\\[-2ex]
 &  & $\mu$ & $\sigma$ &  & $\mu$ & $\sigma$ &  & $\mu$ & $\sigma$ \\ \midrule
STL   &  & 90.19 & 0.01 &  & 88.54 & 0.01 &  & 0.0000 & 0.0083 \\
MTL   &  & 90.28 & \textbf{0.00} &  & 88.56 & 0.01 &  & 0.0612 & \textbf{0.0060} \\
MMOE  &  & 90.29 & 0.02 &  & 88.60 & 0.01 &  & 0.0893 & 0.0162 \\
PLE   &  & 90.30 & 0.01 &  & 88.59 & 0.01 &  & 0.0892 & 0.0074 \\
TabT  &  & 90.16 & 0.01 &  & 88.51 & \textbf{0.01} &  & -0.0336 & 0.0056 \\
FT-T   &  & 90.13 & 0.03 &  & 88.46 & 0.03 &  & -0.0784 & 0.0315 \\
SAINT &  & \textbf{90.31} & 0.03 &  & 88.59 & 0.02 &  & 0.0948 & 0.0109 \\
STEM  &  & 90.28 & 0.02 &  & 88.58 & 0.01 &  & 0.0725 & 0.0106 \\
MultiTab-Net   &  & 90.28 & 0.02 &  & \textbf{88.64} & 0.02 &  & \textbf{0.1064} & 0.0167 \\ \bottomrule
\end{tabular}
\caption{Results on ACS Income, including mean ($\mu$) and standard deviation ($\sigma$) for AUC on the Income (binary) and Marital (multi-class) tasks across 5 seeds.}
\label{tab:inc_extra}
\end{table}

\begin{table}[h!]
\centering
\setlength{\tabcolsep}{3pt}
\resizebox{\textwidth}{!}{%
\begin{tabular}{@{}ccccccccccccccccccc@{}}
\toprule
\multirow{2}{*}{Model} & \multicolumn{2}{c}{Target AUC} & \multicolumn{2}{c}{HLP1 EV} & \multicolumn{2}{c}{HLP2 EV} & \multicolumn{2}{c}{HLP3 EV} & \multicolumn{2}{c}{HLP4 EV} & \multicolumn{2}{c}{HLP5 EV} & \multicolumn{2}{c}{HLP6 EV} & \multicolumn{2}{c}{HLP7 EV} & \multicolumn{2}{c}{$\Delta_m$} \\
\cmidrule(lr){2-3} \cmidrule(lr){4-5} \cmidrule(lr){6-7} \cmidrule(lr){8-9} \cmidrule(lr){10-11} \cmidrule(lr){12-13} \cmidrule(lr){14-15} \cmidrule(lr){16-17} \cmidrule(lr){18-19}\\[-2ex]
 & $\mu$ & $\sigma$ & $\mu$ & $\sigma$ & $\mu$ & $\sigma$ & $\mu$ & $\sigma$ & $\mu$ & $\sigma$ & $\mu$ & $\sigma$ & $\mu$ & $\sigma$ & $\mu$ & $\sigma$ & $\mu$ & $\sigma$ \\
\midrule
STL & 84.90 & 0.11 & 94.39 & 0.09 & 32.42 & 0.08 & 38.79 & 0.24 & 60.43 & \textbf{0.21} & 99.19 & 0.11 & 68.16 & 0.06 & 62.82 & 0.04 & 0.0000 & 0.1555 \\
MTL & 84.13 & 0.46 & 93.72 & 0.17 & 32.66 & 0.27 & 38.60 & 0.20 & 59.34 & 0.51 & 96.67 & 0.48 & 68.39 & 0.15 & 62.93 & 0.13 & -0.6531 & 0.4533 \\
MMoE & 85.03 & 0.14 & 93.98 & \textbf{0.05} & 32.34 & 0.25 & 38.48 & 0.11 & 59.95 & 0.31 & 97.91 & 0.09 & 68.39 & \textbf{0.04} & 62.99 & \textbf{0.06} & -0.3525 & 0.0669 \\
PLE & 85.31 & 0.13 & 94.30 & \textbf{0.05} & 32.37 & 0.25 & 38.84 & \textbf{0.11} & 60.27 & 0.26 & 98.20 & 0.27 & 68.40 & 0.11 & 63.01 & 0.12 & -0.0314 & \textbf{0.0348} \\
TabT & 72.61 & \textbf{0.10} & 78.90 & 0.52 & 26.10 & 0.23 & 28.66 & 0.15 & 37.54 & 0.44 & 57.23 & 1.84 & 54.90 & 0.71 & 48.88 & 0.43 & -24.7917 & 0.1395 \\
FT-T & 78.38 & 0.10 & 94.42 & 0.11 & 31.59 & 0.99 & 36.52 & 0.17 & 54.04 & 3.07 & \textbf{99.87} & \textbf{0.01} & 66.85 & 0.21 & 63.05 & 0.16 & -3.4380 & 0.9582 \\
SAINT & 85.54 & 1.64 & \textbf{94.44} & 0.53 & 31.14 & 1.20 & 37.14 & 1.60 & 57.96 & 5.27 & 98.14 & 2.07 & 67.66 & 0.76 & 62.87 & 0.44 & -1.6514 & 1.5630 \\
STEM & 85.32 & 0.19 & 94.29 & 0.06 & 32.59 & 0.23 & 38.77 & 0.21 & 60.36 & 0.53 & 98.35 & 0.22 & 68.36 & 0.18 & 62.98 & 0.10 & 0.0571 & 0.2218 \\
MultiTab-Net & \textbf{85.99} & 0.24 & 93.99 & 0.11 & \textbf{33.62} & 0.36 & \textbf{39.82} & 0.31 & \textbf{61.36} & 0.30 & 97.96 & 0.19 & \textbf{68.93} & 0.15 & \textbf{63.58} & 0.18 & \textbf{1.2327} & 0.3723 \\
\bottomrule
\end{tabular}}
\caption{Results on Higgs, including mean ($\mu$) and standard deviation ($\sigma$) for AUC on the Target (binary) task and EV on the HLP1–7 (regression) tasks across 5 seeds. Final columns report $\Delta_m$ (mean and std).}
\label{tab:higgs_ev_scaled}
\end{table}

To determine whether MultiTab-Net’s observed improvements are statistically significant, we conducted one-sided independent t-tests on the per-dataset $\Delta_m$ values, comparing MultiTab-Net to each baseline (Table~\ref{tab:t-tests}). This test assumes normally distributed differences between model performances and independence between samples. Across all three datasets, the results demonstrate that MultiTab-Net’s improvements are statistically significant ($p < 0.05$) over all baselines, with three exceptions: PLE on both AliExpress and ACS Income, and SAINT on ACS Income, where the $p$-values did not reach the significance threshold. To complement the t-test and account for small sample sizes and potential non-normality, we also applied a one-sided Wilcoxon signed-rank test (Table~\ref{tab:p-matrix-combined}). This non-parametric test compares the ranks of paired values and makes fewer assumptions about the underlying data distribution. At the dataset level, we used Wilcoxon tests to compare $\Delta_m$ across tasks within each dataset. These results were consistent with the t-test, reaffirming the significance of MultiTab-Net’s improvements in nearly all cases. At the global level, we aggregated $\Delta_m$ values across all datasets and seeds (15 total values per model), and found that MultiTab-Net significantly outperforms all baselines ($p < 0.05$) across the board. This global analysis provides strong evidence for the robustness and generalizability of MultiTab-Net across diverse multitask tabular learning settings.

\begin{table}[h!]
\centering
\setlength{\tabcolsep}{5pt}
\begin{tabular}{@{}ccccccccc@{}}
\toprule
\textbf{Dataset} & \multicolumn{1}{c}{\textbf{STL}} & \multicolumn{1}{c}{\textbf{MTL}} & \multicolumn{1}{c}{\textbf{MMoE}} & \multicolumn{1}{c}{\textbf{PLE}} & \multicolumn{1}{c}{\textbf{TabT}} & \multicolumn{1}{c}{\textbf{FT-T}} & \multicolumn{1}{c}{\textbf{SAINT}} & \multicolumn{1}{c}{\textbf{STEM}}\\ \midrule
AliExpress & \fbox{0.0213} & \fbox{0.0456} & \fbox{0.0026} & 0.1543 & \fbox{0.0114} & \fbox{0.0405} & \fbox{0.0016} & \fbox{0.0262} \\
ACS Income & \fbox{0.0002} & \fbox{0.0035} & 0.1054 & 0.1545 & \fbox{0.0000} & \fbox{0.0002} & 0.1208 & \fbox{0.0011} \\
Higgs & \fbox{0.0003} & \fbox{0.0027} & \fbox{0.0003} & \fbox{0.0008} & \fbox{0.0000} & \fbox{0.0001} & \fbox{0.0431} & \fbox{0.0006} \\ \bottomrule
\end{tabular}
\caption{One-sided t-test p-values comparing MultiTab-Net's $\Delta_m$ to each baseline model on AliExpress, ACS Income, and Higgs datasets. \fbox{Boxed} values indicate statistically significant improvements ($p<0.05$).}
\label{tab:t-tests}
\end{table}

\begin{table}[h!]
\setlength{\tabcolsep}{5pt}
\centering
\begin{tabular}{@{}ccccccccc@{}}
\toprule
\textbf{Dataset} & \textbf{STL} & \textbf{MTL} & \textbf{MMoE} & \textbf{PLE} & \textbf{TabT} & \textbf{FT-T} & \textbf{SAINT} & \textbf{STEM} \\ 
\midrule
AliExpress & \fbox{0.0244} & \fbox{0.0322} & \fbox{0.0137} & 0.1377 & \fbox{0.0420} & \fbox{0.0322} & \fbox{0.0322} & \fbox{0.0186} \\
ACS Income & \fbox{0.0010} & \fbox{0.0244} & 0.1875 & 0.2783 & \fbox{0.0010} & \fbox{0.0010} & 0.3848 & \fbox{0.0244} \\
Higgs & \fbox{0.0007} & \fbox{0.0000} & \fbox{0.0000} & \fbox{0.0000} & \fbox{0.0000} & \fbox{0.0000} & \fbox{0.0008} & \fbox{0.0000} \\
\midrule
\midrule
Global & \fbox{0.0001} & \fbox{0.0006} & \fbox{0.0002} & \fbox{0.0062} & \fbox{0.0000} & \fbox{0.0001} & \fbox{0.0017} & \fbox{0.0000} \\
\bottomrule
\end{tabular}
\caption{One-sided Wilcoxon signed-rank test p-values comparing MultiTab-Net to each baseline model. Each dataset-level comparison is performed across the per-task $\Delta_m$ values (5 seeds per task). The global comparison aggregates $\Delta_m$ across seeds and datasets (15 total points per model). \fbox{Boxed} values indicate statistically significant improvements ($p<0.05$).}
\label{tab:p-matrix-combined}
\end{table}

\subsection{Additional Regression Results on Higgs}
To provide additional transparency for the regression performance of each model, Table~\ref{tab:higgs_mse} reports the mean squared error (MSE) and corresponding standard deviation for each of the seven regression tasks on the Higgs dataset. While our main analysis focuses on explained variance (EV), we recognize that MSE remains a common metric for regression tasks and include it here for reference. Interestingly, the relative rankings of models by task are consistent between EV and MSE. That is, for each individual regression task, MultiTab-Net achieves the same rank compared to other baselines whether performance is measured by EV or MSE. This consistency confirms that our conclusions regarding model performance are not dependent on the choice of regression metric, reinforcing the robustness of MultiTab-Net’s performance. We also note that MultiTab-Net achieves the lowest average MSE across all regression tasks, further supporting its overall effectiveness in minimizing prediction error. That said, we adopt explained variance (EV) rather than MSE as our default metric for calculating $\Delta_m$ in the multitask evaluation. This choice is motivated by several practical considerations. In particular, EV is more robust to outliers and better normalizes across tasks, which is crucial in multitask settings. For example, HLP5 is an easy regression task where all models achieve MSE values near zero. Because $\Delta_m$ is computed using relative changes, small differences in such low-MSE tasks can artificially inflate the contribution to the overall score. A tiny increase in error on such a task could drastically affect $\Delta_m$, even if the difference is practically negligible. In contrast, EV is bounded between 0 and 1, which stabilizes the scale of contributions across tasks. Moreover, using EV makes the range of values comparable to AUC, which is the preferred metric for classification tasks due to its ability to measure class separability. This alignment allows us to compute $\Delta_m$ in a unified and balanced way across both classification and regression tasks, ensuring fair comparisons in a multitask learning setting. Overall, while we include MSE results for completeness, our decision to use EV in the main evaluation stems from both theoretical advantages and empirical stability—particularly in settings with varying task difficulty and distributional characteristics.

\begin{table}[h!]
\centering
\setlength{\tabcolsep}{4pt}
\resizebox{\textwidth}{!}{%
\begin{tabular}{@{}ccccccccccccccccc@{}}
\toprule
\multirow{2}{*}{Model} & \multicolumn{2}{c}{HLP1 MSE} & \multicolumn{2}{c}{HLP2 MSE} & \multicolumn{2}{c}{HLP3 MSE} & \multicolumn{2}{c}{HLP4 MSE} & \multicolumn{2}{c}{HLP5 MSE} & \multicolumn{2}{c}{HLP6 MSE} & \multicolumn{2}{c}{HLP7 MSE} & \multicolumn{1}{c}{Avg. MSE} \\
\cmidrule(lr){2-3} \cmidrule(lr){4-5} \cmidrule(lr){6-7} \cmidrule(lr){8-9} \cmidrule(lr){10-11} \cmidrule(lr){12-13} \cmidrule(lr){14-15}
 & $\mu$ & $\sigma$ & $\mu$ & $\sigma$ & $\mu$ & $\sigma$ & $\mu$ & $\sigma$ & $\mu$ & $\sigma$ & $\mu$ & $\sigma$ & $\mu$ & $\sigma$ &  \\
\midrule
STL & 0.0156 & 0.0003 & 0.2998 & \textbf{0.0002} & 0.0868 & 0.0003 & 0.0627 & \textbf{0.0004} & 0.0002 & 0.0000 & 0.0422 & 0.0002 & 0.0362 & \textbf{0.0000} & 0.0776 \\
MTL & 0.0174 & 0.0005 & 0.2986 & 0.0012 & 0.0871 & 0.0003 & 0.0644 & 0.0008 & 0.0009 & 0.0002 & 0.0418 & 0.0002 & 0.0361 & 0.0001 & 0.0780 \\
MMoE & 0.0168 & 0.0002 & 0.3002 & 0.0010 & 0.0873 & 0.0002 & 0.0635 & 0.0005 & 0.0006 & 0.0000 & 0.0419 & \textbf{0.0001} & 0.0361 & 0.0001 & 0.0780 \\
PLE & 0.0158 & \textbf{0.0001} & 0.3000 & 0.0011 & 0.0867 & \textbf{0.0001} & 0.0630 & 0.0005 & 0.0005 & 0.0001 & 0.0418 & 0.0001 & 0.0361 & 0.0001 & 0.0777 \\
TabT & 0.0586 & 0.0015 & 0.3280 & 0.0011 & 0.1012 & 0.0002 & 0.0989 & 0.0007 & 0.0116 & 0.0005 & 0.0597 & 0.0010 & 0.0499 & 0.0004 & 0.1011 \\
FT-T & 0.0155 & 0.0003 & 0.3036 & 0.0043 & 0.0900 & 0.0003 & 0.0730 & 0.0048 & \textbf{0.0000} & \textbf{0.0000} & 0.0440 & 0.0003 & 0.0360 & 0.0001 & 0.0803 \\
SAINT & \textbf{0.0154} & 0.0015 & 0.3055 & 0.0053 & 0.0892 & 0.0023 & 0.0667 & 0.0085 & 0.0005 & 0.0006 & 0.0428 & 0.0010 & 0.0363 & 0.0005 & 0.0795 \\
STEM & 0.0158 & 0.0002 & 0.2990 & 0.0010 & 0.0870 & 0.0002 & 0.0628 & 0.0008 & 0.0005 & 0.0000 & 0.0419 & 0.0002 & 0.0362 & 0.0001 & 0.0776 \\
MultiTab-Net & 0.0168 & 0.0005 & \textbf{0.2945} & 0.0016 & \textbf{0.0855} & 0.0005 & \textbf{0.0613} & 0.0005 & 0.0006 & 0.0001 & \textbf{0.0411} & 0.0002 & \textbf{0.0355} & 0.0002 & \textbf{0.0765} \\
\bottomrule
\end{tabular}}
\caption{Additional results on Higgs, including mean ($\mu$) and standard deviation ($\sigma$) for MSE on the HLP1–7 regression tasks across 5 seeds. Final column shows the average mean MSE.}
\label{tab:higgs_mse}
\end{table}

\subsection{Statement on Tabular Foundation Models}
Recent work on tabular foundation models, such as TabPFN and TabICL, has proposed using in-context learning as an alternative to traditional gradient-based learning for tabular prediction tasks. These models are designed to generalize across tasks by learning a meta-model trained on a large distribution of synthetic tasks. During inference, they receive the entire training set as input, concatenated along with a query point, and make predictions by conditioning directly on the context of training examples, without further weight updates. This in-context learning paradigm enables zero-shot generalization across tasks, but it introduces a critical scalability limitation: since the model must process all training samples simultaneously in its context window, the number of training examples is fundamentally constrained by the model’s context length and memory footprint. As a result, existing in-context tabular models like TabPFN and TabICL are not applicable to large-scale datasets, such as the ones considered in our benchmark, which contain hundreds of thousands to millions of samples. In fact, current versions of TabPFN explicitly cap the number of training samples at 10,000 samples. While TabPFN does support handling larger datasets using either (1) a tree-based model with node subsampling, or (2) random subsampling of the training data to 10k examples, these strategies are far from optimal in the high-data regime. To empirically test this limitation, we attempted to run the proposed large-dataset variants of TabPFN. However, these attempts consistently timed out or ran out of memory, even when adhering to the recommended 10k sample limit. We then implemented our own workaround by subsampling the training set to 10k and performing inference on the full test set. Despite this, we again encountered out-of-memory errors. We further tried batching the test set in smaller chunks, down to just 500  examples (less than 0.009\% of the AliExpress test set), but even this resulted in out-of-memory crashes. These repeated failures highlight the severe scalability issues that current in-context learning-based tabular models face. As such, while promising in small-scale or few-shot settings, these methods are not yet practical for real-world, large-scale multitask learning, which is the explicit focus of our work.

\subsection{Synthetic Dataset Configuration Details}

\begin{table}[h]
\centering
\setlength{\tabcolsep}{5pt}
\begin{tabular}{ccccc}
\toprule
\multirow{2}{*}{Params.} &  & \multicolumn{3}{c}{\textit{Variation}} \\
\cline{3-5}\\[-2ex]
 &  & \textbf{Correlation} & \textbf{Complexity} & \textbf{Count} \\
 \midrule
$p$ &  & 0.2, 0.6, 1.0 & 0.6 & 0.6 \\
\\[-2ex]
PD &  & {[}3, 3, 3{]} & \begin{tabular}[c]{@{}c@{}}\hspace{2pt}{[}3, 3, 3{]},\\  \hspace{2pt}{[}4, 4, 4{]},\\{[}5, 5, 5{]}\end{tabular} & {[}3, …, 3{]} \\
\\[-2ex]
t &  & 3 & 3 & 3, 5, 7\\
\bottomrule
\end{tabular}
\caption{The parameters used to generate the synthetic datasets with the corresponding results presented in Figure~\ref{fig:syn_variation}.}
\label{tab:gen_params}
\end{table}

To systematically evaluate our multitask models, we tune our synthetic data generator to isolate model performance to variations in task correlation, complexity, and count. The parameters used to generate the datasets can be found in Table~\ref{tab:gen_params}. In general, we have 3 main parameters that govern the multitask dynamics: the pair-wise weight cosine similarity ($p$), the polynomial degree (PD), and the number of tasks (t). In each variation, the corresponding parameter is varied across 3 values, while the other parameters are held fixed. In addition to these parameters, we also set \texttt{num\_features} to 32 and \texttt{num\_samples} to 1,000,000 for all experiments. Finally, to simulate real-world uncertainty in the task labels, we apply task-specific noise sampled from a standard normal distribution that is scaled by a factor of 0.01 (\texttt{noise\_levels}) for all experiments. The PyTorch implementation code for the synthetic dataset generator used in our experiments can be found in Listing~\ref{fig:code}.

\begin{listing}[htb!]
\caption{PyTorch code for synthetic dataset generator used for controlled multitask experiments.}
\begin{lstlisting}[language=Python]
class SyntheticDataset(Dataset):
    def __init__(self, tasks, input_dim, correlation, num_samples, polynomial_degrees, noise_levels=None):
        self.tasks = tasks
        self.num_tasks = len(tasks)
        self.input_dim = input_dim # num_features
        self.correlation = correlation
        self.num_samples = num_samples
        self.pd = polynomial_degrees
        self.noise_levels = noise_levels if noise_levels else [0.01] * self.num_tasks
        self.U = self._generate_orthogonal_unit_vectors()
        self.P = self._construct_gram_matrix()
        self.W = self._construct_weight_matrix()
        self.X = torch.randn(self.num_samples, self.input_dim)
        self.Y = self._generate_labels()

    def _generate_orthogonal_unit_vectors(self):
        U = torch.randn(self.num_tasks, self.input_dim)
        for i in range(self.num_tasks):
            for j in range(i):
                U[i] -= torch.dot(U[i], U[j]) * U[j]
            U[i] /= U[i].norm()
        return U

    def _construct_gram_matrix(self):
        P = torch.full((self.num_tasks, self.num_tasks), self.correlation)
        P.fill_diagonal_(1.0)
        return P

    def _construct_weight_matrix(self):
        eigvals, eigvecs = torch.linalg.eigh(self.P)
        sqrt_eigvals = torch.sqrt(torch.clamp(eigvals, min=0.0))
        W = eigvecs @ torch.diag(sqrt_eigvals) @ self.U
        return W

    def _generate_labels(self):
        Y = []
        for i in range(self.num_tasks):
            linear_term = self.X @ self.W[i]
            poly_term = sum((linear_term ** k) for k in range(2, self.pd[i] + 1))
            noise = torch.randn(self.num_samples) * self.noise_levels[i]
            Y.append(linear_term + poly_term + noise)
        return torch.stack(Y, dim=1)

    def __len__(self):
        return self.num_samples

    def __getitem__(self, idx):
        sample = {
            'features': self.X[idx],
            'targets': {task: self.Y[idx, i] for i, task in enumerate(self.tasks)}
        }
        return sample
\end{lstlisting}
\label{fig:code}
\end{listing}

\end{document}